\pdfoutput=1

\PassOptionsToPackage{dvipsnames,svgnames,x11names,table}{xcolor}
\documentclass[11pt]{article}

\usepackage[final]{acl}

\usepackage{times}
\usepackage{latexsym}
\usepackage[T1]{fontenc}
\usepackage[utf8]{inputenc}
\usepackage{microtype}
\usepackage{silence}
\usepackage{inconsolata}

\usepackage{setspace}
\usepackage[most]{tcolorbox}
\usepackage[colorinlistoftodos]{todonotes}
\usepackage{array}
\usepackage{booktabs}
\usepackage{tabularx}
\usepackage{placeins}
\usepackage[dvipsnames,svgnames,x11names,table]{xcolor}
\usepackage{colortbl}
\usepackage{multirow,makecell}
\usepackage{cleveref}
\usepackage{epigraph}
\usepackage{pifont}
\usepackage{ragged2e}
\usepackage{arydshln}
\usepackage[most]{tcolorbox}
\usepackage{tikz}
\usetikzlibrary{decorations.shapes}
\usepackage{amsfonts}
\usepackage{amssymb}
\usepackage[normalem]{ulem}
\usepackage{hypcap}
\usepackage{capt-of}
\usepackage{fontawesome}
\usepackage{bm}
\usepackage{atbegshi}
\usepackage{hyperref}
\usepackage{nameref}
\usepackage{graphicx}
\usepackage{colortbl}
\usepackage{subfigure}
\usepackage{xspace,mfirstuc,tabulary}
\usepackage{newunicodechar}
\usepackage{atbegshi}
\usepackage{mdframed}
\usepackage{algorithm}
\usepackage[noend]{algpseudocode}
\usepackage{soul}
\usepackage{bbm}
\usepackage{rotating}
\usepackage{tablefootnote}
\usepackage[outline]{contour}
\usepackage{ifthen}
\usepackage{framed}
\usepackage{amsmath}
\usepackage{mathtools}

\definecolor{OliveGreen}{rgb}{0.0, 0.5, 0.0}
\definecolor{ggreen}{HTML}{f0f5ef}
\definecolor{rred}{HTML}{f9efed}
\definecolor{bblue}{HTML}{eef2f9}
\definecolor{light_purple}{HTML}{dfccfa}
\definecolor{gray}{gray}{0.9}
\definecolor{gray2}{gray}{0.7}
\tcbset{on line, 
        boxsep=1pt, left=0pt,right=0pt,top=0pt,bottom=0pt,
        colframe=white,
        colback=light_purple,  
        highlight math style={enhanced}
}
\newtcbox{\greenbox}[1][]{
  on line,
  boxsep=1pt, left=0pt,right=0pt,top=0pt,bottom=0pt,
        colframe=white,
        colback={green!25},  
        highlight math style={enhanced}, #1
}
\newtcbox{\redbox}[1][]{
  on line,
  boxsep=1pt, left=0pt,right=0pt,top=0pt,bottom=0pt,
        colframe=white,
        colback={red!20},  
        highlight math style={enhanced}, #1
}

\newcommand{\wipinrow}[1]{\textcolor{RubineRed}{\textbf{[WIP] #1}}}

\newcolumntype{C}{>{\arraybackslash}X}
\newcommand{\greencheck}{\textcolor{LimeGreen}{\ding{52}}}
\newcommand{\redcross}{\textcolor{Red2}{\ding{56}}}

\newcommand{\ourmethod}{\textsc{LM-Lexicon}\xspace}
\newcommand\coauth{$^\star$}
\newcommand{\correspond}{$^\dag$}
\newcommand\blfootnote[1]{%
  \begingroup
  \renewcommand\thefootnote{}\footnote{#1}%
  \addtocounter{footnote}{-1}%
  \endgroup
}

\usetikzlibrary{shapes} 
\usetikzlibrary{shadows.blur}
\colorlet{myred}{red!85!black!65}
\colorlet{mycyan}{blue!40!cyan!85!black!55}
\colorlet{myblue}{blue!70!cyan!85!black!55}
\colorlet{mydarkblue}{myblue!90!black!90}
\colorlet{mydarkerblue}{myblue!80!black!90}
\colorlet{mydarkestblue}{myblue!70!black!90}
\colorlet{mylightblue}{blue!50!cyan!80!black!45}
\colorlet{mypink}{magenta!95!red!100!black!72}
\colorlet{mypurple}{blue!50!red!90!black!70}
\colorlet{mydarkpurple}{blue!50!red!70!black!70}
\colorlet{mygreen}{green!60!black!75}
\colorlet{mylightgreen}{green!80!red!90!black!50}
\colorlet{myorange}{orange!95!black!70}
\colorlet{mybrown}{brown!80!orange!95!black!80}
\colorlet{myyellow}{yellow!80!orange!95!black!95}
\colorlet{mylightyellow}{yellow!80!orange!95!black!45}

\def\R{1} 
\tikzstyle{pin}=[very thin,line cap=round]
\tikzset{
  slice/.style={fill=#1,draw=#1!80!black,line join=round,line width=0.4,
                blur shadow={shadow blur steps=20,shadow xshift=0.5,
                             shadow yshift=-1,shadow opacity=40}},
  slice/.default=myblue,
  pics/piechart/.style n args={2}{ 
    code={
      \foreach \frac/\name/\col [
        count=\i,
        remember=\angb as \anga (initially #1), 
        evaluate={\angm=\anga-\frac*1.8;  
                  \angb=\anga-\frac*3.6;  
                  \exp=\R*max(0.02,0.2*(1-\frac/100)^8); 
                  \r=\exp+\R*max(0.5,(0.8-\frac/100));} 
      ] in {#2}{ 
        \coordinate (P\i) at (\angm:\exp+\R);
        \draw[slice=\col] 
          (\angm:\exp) --++ (\anga:\R) arc(\anga:\angb:\R) -- cycle;
        \node[white,scale=0.9]
          at (\angm:\r) {\contour{\col!75!black}{\bm{$\frac\mathbf{\%}$}}};
        \node[\col!70!black,anchor=180+\angm,inner sep=4]
          at (P\i) {\bf\bm{$\name$}};
      }
    }
  },
  pics/piechart2/.style n args={2}{ 
    code={
      \foreach \frac/\name/\col [
        remember=\angb as \anga (initially #1), 
        evaluate={\angm=\anga+\frac*1.8;  
                  \angb=\anga+\frac*3.6;  
                  \exp=\R*max(0.02,0.14*(1-\frac/100)^3); 
                  \r=\exp+\R*max(0.5,(1-\frac/100)^2);} 
      ] in {#2}{ 
        \message{^^J frac=\frac: anga=\anga -> angb=\angb}
        \draw[slice=\col] 
          (\angm:\exp) --++ (\anga:\R) arc(\anga:\angb:\R) -- cycle;
        \ifdim \frac pt > 10pt 
          \node[white,align=center]
            at (\angm:\r) {\boldlabel{$\name$}{\frac}};
        \else 
          \node[\col!5!black,align=center,anchor=180+\angm,inner sep=3]
            at (\angm:\exp+\R) {\boldlabel{$\name$}{\frac}};
        \fi
      }
    }
  },
}

\newcommand\boldlabel[2]{\bf\bm{#1}\\[-1]\small\bm{$#2\mathbf{\%}$}}

\newmdenv[
  backgroundcolor=purple!10,
  skipabove=1em,
  skipbelow=0em,
  leftline=true,
  topline=false,
  bottomline=false,
  rightline=false,
  linecolor=purple!88,
  linewidth=4pt
]{rquote}

\newmdenv[
  backgroundcolor=OliveGreen!8,
  skipabove=1em,
  skipbelow=0em,
  leftline=true,
  topline=false,
  bottomline=false,
  rightline=false,
  linecolor=OliveGreen!88,
  linewidth=4pt
]{gquote}

\makeatletter
\newcommand*{\hyperlinkcite}[1]{\hyper@link{cite}{cite.#1}}
\makeatother

\newcommand{\LogoMiddle}{\texorpdfstring{\raisebox{-0.26em}{\includegraphics[height=1.23em]{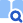}}\xspace}}
\newcommand{\LogoSmall}{\texorpdfstring{\hspace{.1em}\raisebox{-0.26em}{\includegraphics[height=1.05em]{figure/raw/logo.pdf}}\xspace\hspace{-.1em}}}
\newcommand{\GitHubLogo}{\raisebox{-0.15em}{\includegraphics[height=1.3em]{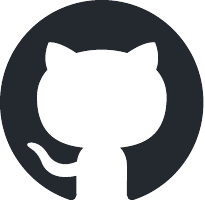}}}
\newcommand{\HFLogo}{\raisebox{-0.28em}{\includegraphics[height=1.2em]{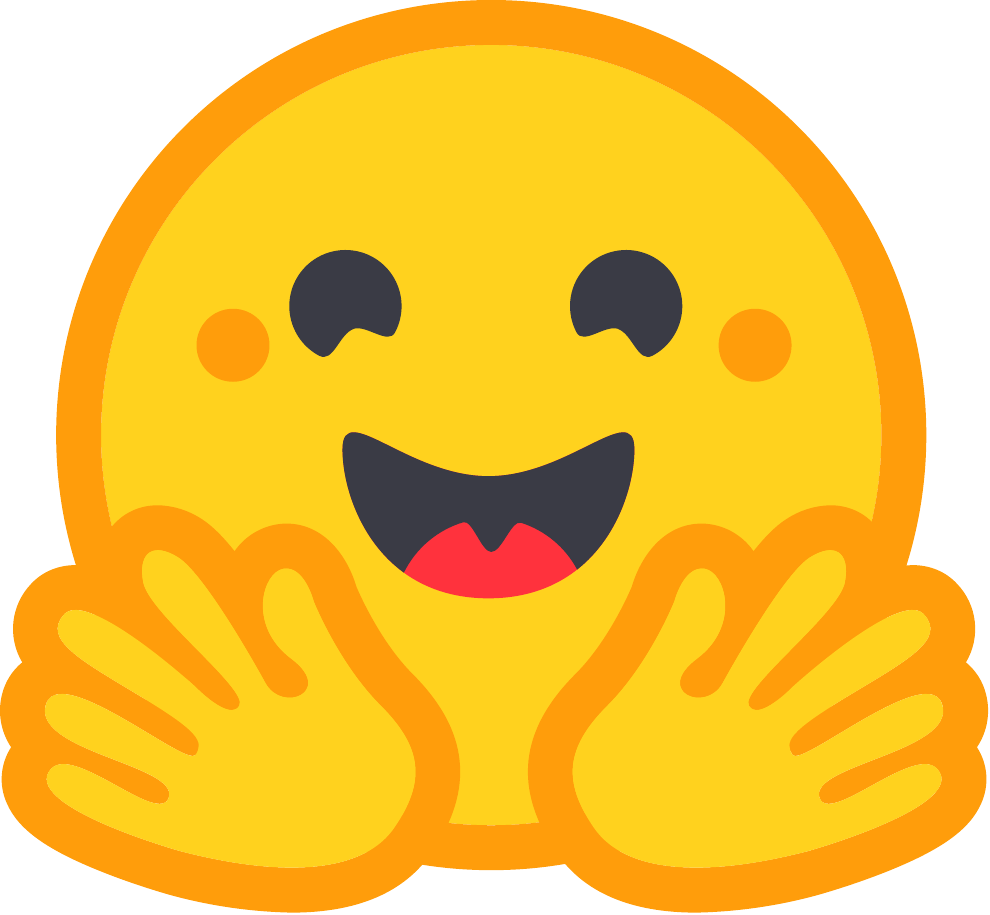}}\xspace}

\definecolor{lightpink}{HTML}{FFDEEF}
\definecolor{lightgreen}{HTML}{E2FEC3}
\definecolor{lightblue}{HTML}{ECF4FF}

\newcommand{\hlcellr}{\cellcolor{BrickRed!5}}
\newcommand{\hlcellrr}{\cellcolor{BrickRed!20}}
\newcommand{\hlcellrrr}{\cellcolor{BrickRed!50}}
\newcommand{\hlcellrrrr}{\cellcolor{BrickRed!70}}

\title{\centering \LogoMiddle\ \textsc{LM-Lexicon:} Improving Definition Modeling\\via Harmonizing Semantic Experts}

\author{Yang Liu$^{1}$\coauth{}\correspond{} \hspace{1.2cm} Jiaye Yang$^{2}$\coauth{} \hspace{1.2cm} Weikang Li$^{3}$ \\ \hspace{1.2cm} \textbf{Jiahui Liang$^{2}$ \hspace{1.2cm} Yang Li$^{2}$ \hspace{1.2cm} Lingyong Yan$^{2}$ \hspace{1.6cm}} \\
  $^{1}$BIGAI \hspace{2.0cm} $^{2}$Baidu Inc. \hspace{1.0cm} $^{3}$Peking University\\
   \texttt{\href{mailto:liuyang@bigai.ai}{liuyang@bigai.ai}} \\\vspace{1em}\GitHubLogo~\url{https://lm-lexicon.github.io}\vspace{-1em}\\\HFLogo~\url{https://huggingface.co/LM-Lexicon}
}

\begin{document}
\maketitle

\begin{abstract}
We introduce \LogoSmall\ \ourmethod, an innovative definition modeling approach that incorporates data clustering, semantic experts learning, and model merging using a sparse mixture-of-experts architecture. By decomposing the definition modeling task into specialized semantic domains, where small language models are trained as domain experts, \LogoSmall\ \ourmethod achieves substantial improvements (+7\% BLEU score compared with the prior state-of-the-art model) over existing methods on five widely used benchmarks. Empirically, we demonstrate that 1) the clustering strategy enables fine-grained expert specialization with nearly 10\% improvement in definition quality; 2) the semantic-aware domain-level routing mechanism achieves higher expert efficacy (+1\%) than conventional token-level routing; and 3) further performance gains can be obtained through test-time compute and semantic expert scaling. Our work advances definition modeling while providing insights into the development of efficient language models for semantic-intensive applications. \blfootnote{\coauth Equal contribution.}\blfootnote{\correspond Correspondence to: liuyang@bigai.ai}
\end{abstract}

\section{Introduction}
\label{sec:introduction}

\begin{figure}[t]
    \centering
    \includegraphics[width=0.9\linewidth]{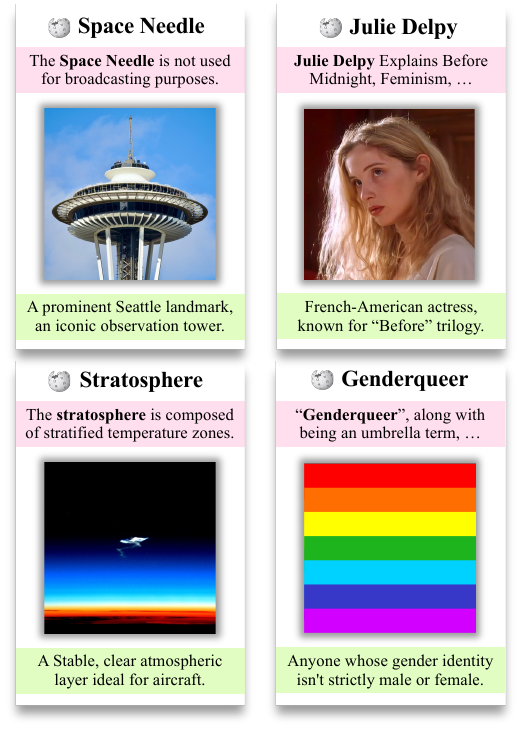}
  \vspace{-0.6em}
  \caption{Four examples of the \textbf{term}, \colorbox{lightpink!80}{context} (input), and \colorbox{lightgreen!80}{definition} (output) for definition modeling task.}
  \label{fig:main-figure}
\end{figure}
Defining terms (Figure \ref{fig:main-figure}) is the first step toward constructing a lexicon for a language \cite{PUSTEJOVSKY1993193}. Precise definitions should be formed as summarized and human-readable sentences that capture the main sense of a term. Modern language use demands continuous updates to include new terms, novel senses, meaning shifts, and domain knowledge \cite{hogeweg2020nature}, yet traditional lexicon construction remains labor-intensive \cite{ahlswede-1985-tool}. To address this challenge, definition modeling (DM) has emerged as a promising approach, where definitions are automatically generated based on the target term and its context \cite[\textit{inter alia}]{giulianelli-etal-2023-interpretable}.

 While existing DM approaches yield reasonable results, they face several key limitations. First, current methods struggle to capture subtle and rare word senses, resulting in incomplete semantic coverage \cite{huang-etal-2021-definition,giulianelli-etal-2023-interpretable,periti-etal-2024-automatically}. Second, even frontier large language models (LLMs), despite their strong language understanding capabilities, tend to generate definitions that are either overly generic or excessively specific \cite{jhirad2023evaluating,yin2023word,almeman-etal-2024-wordnet}. Third, existing methods often fail to handle terms that exhibit different meanings across domains (\textit{e.g.}, technical vs. general usage), a phenomenon known as \textit{semantic heterogeneity} \cite{huang-etal-2021-definition}. Recent attempts such as domain adaptation \cite{zhang2022fine} or multi-task learning \cite{kong-etal-2022-multitasking} have shown limited success. These challenges point to an inherent bottleneck in current LLMs: their dense architectures force polysemantic representation to highly share the same neurons (\textit{i.e.}, superposition) \cite{elhage2022superposition}, making it difficult to maintain precise, domain-specific meaning representations \cite{bricken2023monosemanticity}. Due to the lack of sparsification mechanisms, this architectural constraint affects their ability to generate accurate definitions when words have distinct meanings across different domains.

\begin{figure*}
    \centering
    \includegraphics[width=0.95\linewidth]{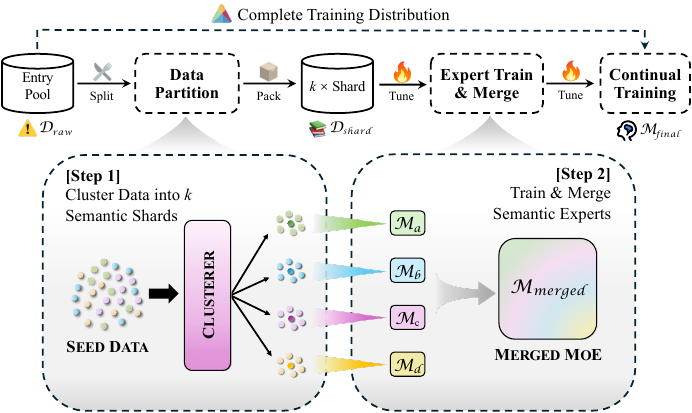}
    \caption{Diagram of \ourmethod (\textit{i.e.}, \textit{Specialize-then-Synthesize}) framework.}
    \label{fig:method}
\end{figure*}


To mitigate these issues, we propose \LogoSmall\ \ourmethod (\textbf{L}anguage \textbf{M}odel as \textbf{Lexicon}), which learns to perform DM covering multiple domains, adapting diverse definition genres with a scalable mixture-of-experts (MoE) architecture. Unlike prior work, such as BTX \cite{sukhbaatar2024branch} and \textsc{Llama-MoE} \cite{zhu-etal-2024-llama}, \textbf{our method incorporates data clustering, semantic expert-specialized MoE, and domain-level sequence routing, obtaining impressive performance gains in DM benchmarks.} As depicted in Figure \ref{fig:method}, instead of training directly on raw definition corpora, our method trains multiple semantic experts parallely, merges them by composing their weights, and routes test samples with the introduced semantic-aware router during inference.

Our contributions can be summarized as follows:
\begin{itemize}
    \item We propose \LogoSmall\ \ \ourmethod, a framework for definition modeling by harmonizing inherent heterogeneity in lexical semantics. It allows specialized semantic experts to be integrated for domain updates, enabling generalization to new domains, or collapsing back to a single expert for efficient inference.
    \item We design a domain-level sequence routing policy in \LogoSmall\ \ourmethod. This method routes representation of samples informed by fine-grained information via semantic domains identified with pre-hoc auto clustering. 
    \item Extensive experiments across five benchmarks validate the effectiveness of \LogoSmall\ \ourmethod. Notably, in automatic evaluation, \LogoSmall\ \ourmethod shows up to 10\% improvement over strong baselines. Furthermore, \LogoSmall\ \ \ourmethod excels across most criteria in human evaluation, particularly outperforming frontier LLMs in semantic-intensive scenarios, where even many-shot setups fail to produce appropriate definitions. 
\end{itemize}
\section{Related Work}
\label{sec:related-work}

\paragraph{Upcycling to Mixture-of-Experts.}
On the aspect of model efficiency and expressiveness, \citet{fedus2022switch,jiang2024mixtral,Shao2024DeepSeekV2AS} focus on designing efficient MoE architecture with token-level router. From the expert specialization aspect, \citet{li2022branch} introduced Branch-Train-Merge (BTM) that learns expert LMs specialized to different domains and \citet{sukhbaatar2024branch} developed Branch-Train-MiX (BTX), which composes a set of specialized LMs by their feed-forward networks. In addition, \citet{zoph2022st,jiang2024mixtral,petridis2024constitutionalexperts,ma2024mode} revealed the efficacy of expert specialization at the lexicon, structured syntactic, and semantic domain level, respectively. However, these works adopt conventional routing schemes, such as token-level TopK routing, rather than exploring those better suited for semantic-intensive scenarios.

\paragraph{Definition Modeling.} 
Several early studies on DM \cite[\textit{inter alia}]{noraset2017definition,ni2017learning,gadetsky2018conditional,ishiwatari-etal-2019-learning} leveraged pre-trained word embeddings as global or local contexts of a term, to generate definitions of the given target word. Then \citet{huang-etal-2021-definition,kong-etal-2022-multitasking,zhang2022fine,giulianelli-etal-2023-interpretable,periti-etal-2024-automatically} propose methods for DM using Transformer-based Seq2Seq LMs (\textit{e.g.}, T5) and Causal LMs. In the era of LLM, \citet{jhirad2023evaluating} and \citet{yin2023word} used LLMs such as GPT-3.5 and GPT-4 to perform DM with in-context learning tailored to diverse domains. \citet{periti-etal-2024-automatically} explored training causal LMs to generate definitions with instruction tuning; however, they still lack a detailed quality evaluation and comphrehensive comparison with baselines.
\section{Methodology}
\label{sec:lm-lexicon}
In this section, we present the details of our proposed \LogoSmall\ \ourmethod framework. \S\ref{subsec:overview_of_method} introduces the formulation to illustrate the main idea. In \S\ref{subsec:learning_domain_semantic_experts}, we illustrate the design of semantic expert specialization, followed by model merging in \S\ref{subsec:merging_experts_into_a_unified_moe}. 

\subsection{Overview of \LogoSmall\ \ \ourmethod}
\label{subsec:overview_of_method}
Given a seed model $\mathcal{M}$ that has been pre-trained, our goal is to improve its multi-domain performance in lexical semantics. As shown in Fig. \ref{fig:method}, the framework of \LogoSmall\ \ourmethod consists of two components: \textbf{(1) semantic expert specialization} and \textbf{(2) MoE model merging}. The proposed method contains three stages, training data partitioning, parallel expert training, and separate experts merging, \textit{i.e.}, the \textit{Specialize-then-Synthesize} framework. Considering the heterogeneity of glosses, we split the training data into semantically distinctive clusters to facilitate expert learning. To model various domains, we use separate models to learn domain-specific knowledge asynchronously. To perform the DM task generally, we merge these experts into a single MoE model for further fine-tuning.

\subsection{Learning Domain-specific Semantic Experts}
\label{subsec:learning_domain_semantic_experts}
\paragraph{Dataset Construction.} Training data $\mathcal{D}$ consists of triplets $\langle c,t,d \rangle$, where $c$ represents the context in which the term is used (either a sentence or phrase), $t$ denotes the term itself, and $d$ is its reference definition. A concatenated sequence is then formatted using the prompt template $p(\cdot, \cdot)$ as input. Specifically, we follow \citet{giulianelli-etal-2023-interpretable} to use $p\coloneqq$ \textsc{<bos>``$\{\{c\}\}$'' What is the definition of ``$\{\{t\}\}$''<eos>} as the prompt template.

\paragraph{Clustering.}
\LogoSmall\ \ourmethod begins with the training data partitioning since merging without it could lead to a group of homogeneous experts. To cluster training data, we calculate the embeddings of $p(c,t)$ in each training sample with \textit{nvidia-embed-v2} \cite{lee2025nvembedimprovedtechniquestraining}, and then cluster with balanced \textit{k}-means \cite{balancedkmeans}. This process results in $N$ clusters in terms of lexical semantics, each related to a semantic domain such as adjectives and proper nouns (see Fig. \ref{fig:3d-ex-visualized-cluster}), corresponding to partitioned training datasets $\mathcal{D} \coloneqq \{\mathcal{D}_1, \ldots, \mathcal{D}_N \}$. It also produces $N$ cluster centroids $\{v_1, v_2, \ldots, v_n\}$. In the present study, we perform pre-experiments to determine the number of clusters and select $N=4$ as the best cluster numbers by the cluster cohesion and separation in the DM scenario (See Appendix \S\ref{subsec:data_clustering_results}), as well as considering the training and inference efficiency.

\paragraph{Experts Training.} 
Initializing from a seed model $\mathcal{M}$, we train $N$ $\times$ LMs: $\{ \mathcal{M}_1, \ldots, \mathcal{M}_N \}$ as experts, with each model $\mathcal{M}_i$ being trained on the corresponding dataset $\mathcal{D}_i$, using the negative loglikelihood (NLL) loss in Eq. \ref{eq:loss}:
\begin{equation}
\begin{aligned}\label{eq:loss}
    \mathcal{L}_{\text{NLL}} = &- \mathbb{E}_{(c, t, d) \sim \mathcal{D}}\left[\log\mathcal{P}(\hat{d}\mid p(c,t))\right] .
\end{aligned}
\end{equation}
Here, $\hat{d}$ denotes the definition predicted by the model, given the prompt $p(\cdot, \cdot)$. We employ a loss-masking strategy to omit the tokens of prompt during loss computation, ensuring that gradients are only propagated through tokens in the part of predicted definition. When expert training finished, we end up with $N$ different LMs, with each specialized in a domain $\mathcal{D}_i$.

\subsection{Merging Experts into a Unified MoE}
\label{subsec:merging_experts_into_a_unified_moe}
After all domain experts are obtained, previous works either average the final output distributions of experts to generate next token \cite{gururangan2023scaling} or select experts by determining which domain the input belongs to at the test time \cite{li2022branch}. Differently, we perform MoE Upcycling by merging the weights of experts, aiming at mixing model capabilities across diverse domains.

\paragraph{Model Merging.} We combine semantic experts into a unified MoE to exploit the parametric domain capability \cite{sukhbaatar2024branch,zhou-etal-2025-mergeme}. In the composition, \LogoSmall\ \ourmethod brings together the feed-forward networks (FFNs) of the expert models as expert layers in MoE and averages the remaining parameters. Specifically, if $\text{FFN}_i^\ell(x)$ is the FFNs at the $\ell$-th layer of the $i$-th expert $\mathcal{M}_i$, then the combined MoE layer for input representation $x$ at layer $\ell$ will be computed as:
\begin{equation}
\begin{aligned}\label{eq:moe_routing}
    \text{FFN}_\text{MoE}^\ell(x) = \sum_{i=1}^N \mathcal{G}(x)\cdot\text{FFN}_i^\ell(x).
\end{aligned}
\end{equation}
where $\mathcal{G}(\cdot)$ is a semantic domain-level router. During both training and inference, the input representation $x$ will be routed to the nearest centroid by computing its pairwise cosine similarity with each semantic label (\textit{i.e.}, the centroid of a domain cluster), as illustrated in \S\ref{subsec:learning_domain_semantic_experts}. $\mathcal{G}(\cdot)$ usually has a sparse output and hence switches on only some experts. In \LogoSmall\ \ourmethod, we start from top-k (k = 2) routing \cite{shazeer2017}, where $\mathcal{G}(x) = \text{Softmax}(\text{TopK}(W^\ell x))$, where $W^\ell$ is a linear transformation in router. For multihead self-attention (MHA) sublayers and the remaining parameters (\textit{e.g.,} embedding layer), we average the weights of domains.
 The merging process of MoE model is provided in Algorithm \ref{alg:algo1}.

The above merging model into a MoE introduces router $\mathcal{G}$ with new parameters $W^\ell$, which requires further learning to make optimal choices. To enhance semantic-aware experts after merging, we continue to slightly fine-tune the router $\mathcal{G}$ and selected expert layers to coordinate them in the semantic representation space \cite{bai-etal-2025-understanding}.

\begin{algorithm}
\caption{Compose MHA and MLP modules for each decoder layer $\ell$ in \LogoSmall\ \ \ourmethod.}\label{alg:algo1}
\begin{algorithmic}[1]
\Require Domain Experts $\mathcal{E} := \{e_1, e_2, \dots, e_n\}$. 
\Ensure  \textsc{LM-Lexicon-MoE ($\mathcal{M}$)}
\Procedure{Modules-Composer}{$\mathcal{E}$}
\State $\mathcal{M}$ $\leftarrow \emptyset$ \Comment{\colorbox{green!20}{\textsc{init state dict}}}
\For{$e_i \in \mathcal{E}$} \Comment{\colorbox{red!20}{\textsc{iterate\ each\ expert}}}
\State $i \leftarrow$ GetExpertIdx($e_i$)
\State \textcolor{gray2}{/* Retrieve MHA and MLP weights */}
\State $\theta_{mha}, \theta_{mlp} \leftarrow$ HookWeights($e_i$)
\For{$\theta \in \{\theta_{mha},\ \theta_{mlp}\}$}
        \If{$\text{IsRouterLayer}(\theta)$}
            \State \textcolor{gray2}{/* Get formatted layer name */}
            \State $n \leftarrow \text{FormatName}(\theta, i)$
            \State $\mathcal{M}[n] \leftarrow \theta$
        \Else \Comment{\colorbox{SkyBlue!50}{\textsc{Average} $\theta$ \textsc{of module}}}
            \State $\mathcal{M}[n] \leftarrow \mathcal{M}.\text{get}(n, \mathbf{0}) + \theta/|\mathcal{E}|$
        \EndIf
    \EndFor
\EndFor
\State \textbf{return} $\mathcal{M}$
\EndProcedure
\end{algorithmic}
\end{algorithm}

\section{Experiments}
\label{sec:experimental-setup}

\subsection{Implementation Details}

\paragraph{Datasets.}
\begin{table*}[t]
    \small
    \centering
    \resizebox{\linewidth}{!}{
	\begin{tabular}{lccccc}
	    \toprule
	    & \textbf{WordNet} & \textbf{Oxford} & \textbf{Wikipedia} & \textbf{Urban} & \textbf{3D-EX} \\
	    \midrule
	    genre & formal & formal & web & idiom & misc.  \\
	    domain & synset & lexicon & encyclopedia & slang & multi \\
	    publish year & 2017 & 2018 & 2018 & 2017 & 2023 \\
	    \midrule
	    \# $\mathcal{S}_\text{train}^t$ & $13,883$ & $97,855$ & $887,455$ & $411,384$ & $1,309,312$ \\
	    \rule{0pt}{2.1ex}\# $\mathcal{S}_\text{valid}^t$ & $1,752$ & $12,232$ & $44,003$ & $57,883$ & $513,789$\\
        \rule{0pt}{2.1ex}\# $\mathcal{S}_\text{test}^t$ & $1,775$ & $12,232$ & $57,232$ & $36,450$ & $450,078$ \\
        \midrule
        \# glo. per term & $1.75\pm1.19$& $2.99\pm4.41$ & $5.86 \pm 78.25 $ & $2.11 \pm 2.92$ & $6.00 \pm 53.78$ \\
        \# tok. per term & $1.00 \pm 0.00 $ & $1.00 \pm 0.00$ & $1.85 \pm 0.93$ & $1.44 \pm 0.72 $ & $1.45 \pm 0.78$ \\
        \# tok. per ctx. & $5.79 \pm 3.44 $ & $19.02 \pm 9.18$ & $19.68 \pm 6.31 $ & $11.36 \pm 6.02 $ & $18.82 \pm 9.99$ \\
        \# tok. per glo. & $6.64 \pm 3.78 $ & $11.41 \pm 7.13$ & $5.97 \pm 4.51 $ & $11.02 \pm 6.86 $ & $8.97 \pm 6.76$ \\
        \% overlap rate & $\greenbox{0.00} / \redbox{\footnotesize{0.00}}$ & $\greenbox{80.72} / \redbox{\footnotesize{0.09}}$ & $\greenbox{0.00} / \redbox{\footnotesize{0.00}}$ & $\greenbox{20.62} / \redbox{\footnotesize{20.56}}$ & $\greenbox{0.00} / \redbox{\footnotesize{0.00}}$ \\
		\bottomrule
	\end{tabular}
    }
	\caption{For datasets used in this paper, we report the mean and standard deviation of per-term, per-context, and per-gloss statistics. We report the number of terms of samples denoted $\mathcal{S}_*^t$ for train, valid, and test splits in each dataset. The lexical overlap of each dataset is computed with $\lvert \mathcal{S}_\text{train}^t \cap\ \mathcal{S}_\text{test}^t \rvert\ /\ \lvert \mathcal{S}_\text{test}^t\rvert$. Specifically, the \greenbox{\%} is computed by intersection rate of term occurrence and the \redbox{\%} is computed by intersection rate of pair-wise ``term $\oplus$ gloss''.}
	\label{tab:datastats}
\end{table*}
We use the benchmarks introduced in \citet{ishiwatari-etal-2019-learning}(see Table \ref{tab:datastats}), which consist of four small datasets and 3D-EX from \citet{almeman-etal-2023-3d} (see details in \S\ref{sec:additional-experiment-results}).    
\begin{itemize}
	\item \textbf{WordNet} \cite{noraset2017definition} is an online dataset\footnote{\url{https://wordnet.princeton.edu}} of terms, definitions, and examples.
	\item \textbf{Oxford} \cite{gadetsky2018conditional} is built on the widely used online oxford dictionary\footnote{\url{https://en.oxforddictionaries.com}}.
	\item \textbf{Wikipedia}\footnote{\url{https://www.wikidata.org}} \cite{ishiwatari-etal-2019-learning} is introduced to test the model capacity on the description of phrases, rather than words.
	\item \textbf{Urban} \cite{ni2017learning}\footnote{\url{https://www.urbandictionary.com}} contains terms of internet slang and urban words.
	\item \textbf{3D-EX} \cite{almeman-etal-2023-3d} is the largest English definition modeling dataset\footnote{\url{https://github.com/F-Almeman/3D-EX}} which comprises many well-known DM resources, including the four mentioned datasets.
\end{itemize}
Note that we perform clustering only on 3D-EX and use the resulting four clusters for finetuning and merging semantic experts.

\paragraph{Compared Baselines.} Llama-3-8B \cite{dubey2024llama} is used as the seed model for asynchronous expert training. We select three categories of strong baseline methods for comparison purposes.
\begin{itemize}
	\item \textbf{Supervised Seq2seq LM:} We reproduce Rerank-T5 \cite{huang-etal-2021-definition}, Contrast-T5 \cite{zhang2022fine}, SimpDefiner \cite{kong-etal-2022-multitasking}, MDM-T5 \cite{zhang2023exploiting}, and Flan-T5-Def \cite{giulianelli-etal-2023-interpretable}.
	\item \textbf{Supervised Causal LM:} We report the in-distribution results of LlamaDictionary \cite{periti-etal-2024-automatically}, which is finetuned on \textbf{\textit{Llama-3-8B-Instruct}}, and assess its out-of-distribution performance for the unseen domains. 
	\item \textbf{Frontier Causal LM:} We test GPT-4-Turbo \cite{achiam2023gpt}, Gemini-1.5-Pro \cite{reid2024gemini}, and Claude-3-Opus \cite{anthropic2024claude} with random exemplar selection (Random-ICL) and retrieval-based exemplar ranking (Retrieval-ICL) based on \citet{wu-etal-2023-openicl} in many-shot settings.
\end{itemize}

\begin{figure}[t]
  \centering
  \resizebox{1.05\linewidth}{!}{\includegraphics{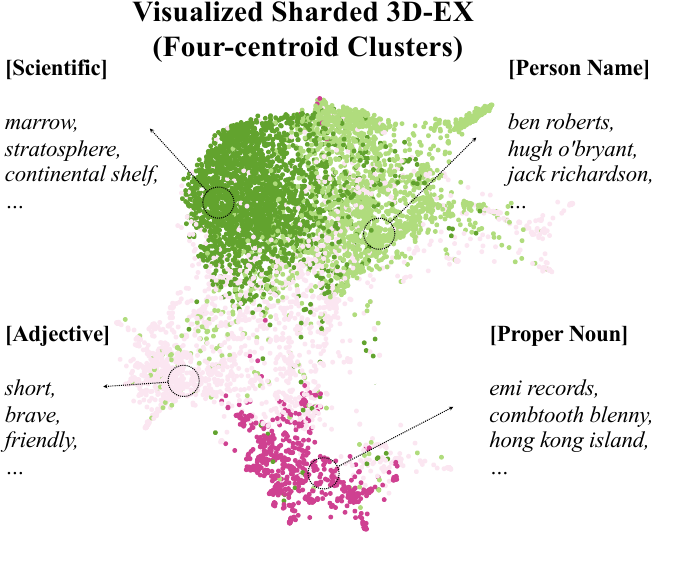}}
  \caption{Four-cluster UMAP plot of 10K random definitions of terms in 3D-EX (\S\ref{sec:experimental-setup}). Each cluster is assigned manually with a \textbf{[label]} by their major constituents.}
  \label{fig:3d-ex-visualized-cluster}
  \vspace{-1.3em}
\end{figure}

\paragraph{Training and Evaluation Details.}
We run instruction tuning on four clusters obtained from 3D-EX respectively. The models trained on four clusters of 3D-EX are merged through \S\ref{subsec:merging_experts_into_a_unified_moe}. After merging, we proceed to fine-tune the MoE model to learn routers using the full 3D-EX dataset. In addition, we perform instruction tuning on the four real-world datasets. The hyperparameters can be found in the Tab. \ref{tab:training-hyperparams}. We run three times with seeds to report the mean results and the standard deviation, with seed $s_i \in \left\{21, 42, 84\right\}$.
All experiments are conducted on 8 $\times$ NVIDIA H100. Model sizes and training FLOPs are reported in Table \ref{tab:carbon-footprint}. 

We employ metrics including (1) lexical n-gram-based: \textsc{Bleu} \cite{papineni-etal-2002-bleu}, \textsc{Rouge-L} \cite{lin-2004-rouge}, and \textsc{Meteor} \cite{lavie-agarwal-2007-meteor}; (2) semantic-based: \textsc{BertScore} \cite{zhang2019bertscore}, \textsc{MoverScore} \cite{zhao2019moverscore}, and \textsc{Mauve} \cite{pillutla2021mauve}. We reuse the implementation of \textsc{Bleu} in \citet{huang-etal-2021-definition}, \textsc{Rouge} and \textsc{BertScore} used in \citet{giulianelli-etal-2023-interpretable}, as well as the rest of metrics for evaluation. To further evaluate the effectiveness of our method, we perform a human evaluation described in \S\ref{subsec:main_results}.

\subsection{Main Results}\label{subsec:main_results}
\begin{table*}[!t]
\begin{center}
\centering
\resizebox{2.0\columnwidth}{!}
{
\begin{tabular}{lcccccccccc>{\columncolor{white}}c}
\toprule
      & \multicolumn{2}{c}{\textbf{WordNet}} & \multicolumn{2}{c}{\textbf{Oxford}} & \multicolumn{2}{c}{\textbf{Wiki}} & \multicolumn{2}{c}{\textbf{Urban}} & \multicolumn{2}{c}{\textbf{3D-EX}} & \cellcolor{white}\\
\cmidrule(lr){2-3}
\cmidrule(lr){4-5}
\cmidrule(lr){6-7}
\cmidrule(lr){8-9}
\cmidrule(lr){10-11}
& BLEU & ROUGE & BLEU & ROUGE &BLEU & ROUGE & BLEU & ROUGE & BLEU & ROUGE & \multirow{-2}{*}{\cellcolor{white}
\multirowcell{-1.7}{\textbf{Avg.} \\ \textbf{Results}} }\\

\midrule

\hyperlinkcite{huang-etal-2021-definition}{Rerank-T5 (2021)}$^\clubsuit$ & $30.91$ & $30.99$ & $25.56$ & $28.00$ & $55.61$ & $57.25$ & $17.77$ & $18.25$ & $34.43$ & $38.57$ & $32.85$ / $34.61$ \\
\hyperlinkcite{zhang2022fine}{Contrast-T5 (2022)}$^\clubsuit$ & $30.81$ & $26.27$ & $22.51$ & $28.18$ & $55.26$ & $42.27$ & $17.53$ & $16.34$ & $34.27$ & $37.62$ & $32.07$ / $30.13$ \\
\hyperlinkcite{kong-etal-2022-multitasking}{SimpDefiner (2022)}$^\clubsuit$ & $28.91$ & $20.47$ & $23.48$ & $29.59$ & $44.03$ & $49.26$ & $13.54$ & $15.37$ & $32.08$ & $31.57$ & $28.40$ / $29.25$ \\
\hyperlinkcite{zhang2023exploiting}{MDM-T5 (2023)}$^\clubsuit$ & $31.18$ & $32.55$ & $24.16$ & $27.68$ & $54.33$ & $55.83$ & $17.53$ & $17.18$ & $32.67$ & $32.38$ & $31.97$ / $33.12$ \\
\hyperlinkcite{giulianelli-etal-2023-interpretable}{Flan-T5-Def (2023)}$^\clubsuit$ & $31.96$ & $40.45$ & $21.34$ & $32.39$ & $13.82$ & $23.97$ & $5.33$ & $10.61$ & $26.43$ & $25.12$ & $19.77$ / $26.50$ \\
\hyperlinkcite{periti-etal-2024-automatically}{LlamaDict (2024)}$^\clubsuit$ & $33.86$ & $\textbf{43.50}$ & $22.77$ & $\underline{36.46}$ & $14.38$ & $25.29$ & $15.70$ & $14.51$ & $24.56$ & $26.11$ & $22.50$ / $29.17$ \\

\midrule

\multicolumn{12}{l}{\textsc{GPT-4-Turbo}} \\
~  \hspace{-0.2cm}$\hookrightarrow \text{+}$ \textit{Random-ICL} & $30.95$ & $32.61$ & $21.93$ & $30.82$ & $31.63$ & $45.89$ & $11.08$ & $12.19$ & $25.93$ & $34.48$ & $24.30$ / $31.19$ \\
~ \hspace{-0.2cm}$\hookrightarrow \text{+}$ \textit{Retrieval-ICL} & $27.46$ & $29.74$ & $20.44$ & $34.35$ & $35.40$ & $40.68$ & $22.53$ & $26.53$ & $29.73$ & $37.66$ & $27.11$ / $33.79$ \\

\multicolumn{12}{l}{\textsc{Claude-3-Opus}} \\
~ \hspace{-0.2cm}$\hookrightarrow \text{+}$ \textit{Random-ICL} & $28.63$ & $27.84$ & $19.99$ & $34.21$ & $23.30$ & $35.22$ & $1.59$ & $3.08$ & $18.57$ & $28.49$ & $18.41$ / $25.76$ \\
~ \hspace{-0.2cm}$\hookrightarrow \text{+}$ \textit{Retrieval-ICL} & $18.57$ & $21.76$ & $15.51$ & $25.99$ & $14.59$ & $15.83$ & $5.93$ & $7.19$ & $17.46$ & $24.67$ & $14.41$ / $19.08$ \\

\multicolumn{12}{l}{\textsc{Gemini-1.5-Pro}} \\
 ~ \hspace{-0.2cm}$\hookrightarrow \text{+}$ \textit{Random-ICL} & $23.42$ & $26.27$ & $25.51$ & $35.97$ & $36.87$ & $48.13$ & $8.44$ & $9.59$ & $29.4$ & $38.02$ & $24.72$ / $31.59$ \\
~ \hspace{-0.2cm}$\hookrightarrow \text{+}$ \textit{Retrieval-ICL} & $25.24$ & $27.88$ & $\textbf{28.10}$ & $\textbf{36.98}$ & $35.59$ & $43.71$ & $8.85$ & $9.18$ & $32.99$ & $39.14$ & $26.15$ / $31.37$ \\

\midrule

\multicolumn{12}{l}{\LogoMiddle\ \textsc{LM-Lexicon-Dense (8B)}} \\
~ \hspace{-0.2cm}$\hookrightarrow \text{+}$ \textit{Zero-shot} & $36.99_{\hspace{0.05cm}0.59}^{*}$ & $37.83_{\hspace{0.05cm}0.45}^{*}$ & $26.09_{\hspace{0.05cm}0.60}$ & $34.55_{\hspace{0.05cm}0.57}^{*}$ & $\text{57.9}^{*}_{\hspace{0.05cm}2.44}$ & $\textbf{59.56}^{*}_{\hspace{0.05cm}1.50}$ & $\underline{26.09}_{\hspace{0.05cm}0.27}^{*}$ & $\underline{28.35}_{\hspace{0.05cm}0.28}^{*}$ & $\text{35.01}^{*}_{\hspace{0.05cm}0.22}$ & $\text{43.32}^{*}_{\hspace{0.05cm}0.27}$ & $\text{34.63}^{*}$ / $\text{38.79}^{*}$ \\
~ \hspace{-0.2cm}$\hookrightarrow \text{+}$ \textit{BoN-Oracle}$^{\dagger}$ & \textcolor{Gray!75}{$\text{47.90}_{\hspace{0.05cm}0.30}$} & \textcolor{Gray!75}{$\text{44.19}_{\hspace{0.05cm}0.80}$} & \textcolor{Gray!75}{$\text{30.07}_{\hspace{0.05cm}0.06}$} & \textcolor{Gray!75}{$\text{42.78}_{\hspace{0.05cm}0.11}$} & \textcolor{Gray!75}{$\text{62.07}_{\hspace{0.05cm}0.11}$} & \textcolor{Gray!75}{$\text{68.62}_{\hspace{0.05cm}0.19}$} & \textcolor{Gray!75}{$\text{36.16}_{\hspace{0.05cm}0.69}$} & \textcolor{Gray!75}{$\text{38.87}_{\hspace{0.05cm}0.47}$} & \textcolor{Gray!75}{$\text{48.78}_{\hspace{0.05cm}0.89}$} & \textcolor{Gray!75}{$\text{49.71}_{\hspace{0.05cm}2.21}$} & \textcolor{Gray!75}{$\text{44.99}$ / $\text{48.83}$} \\
~ \hspace{-0.2cm}$\hookrightarrow \text{+}$ \textit{BoN-ORM} & $\text{37.73}_{\hspace{0.05cm}0.26}^{*}$ & $\text{37.94}_{\hspace{0.05cm}0.38}^{*}$ & $\underline{26.74}_{\hspace{0.05cm}0.18}^{*}$ & $\text{35.18}_{\hspace{0.05cm}0.59}^{*}$ & $\text{59.33}_{\hspace{0.05cm}0.12}^{*}$ & $\underline{59.46}_{\hspace{0.05cm}0.37}^{*}$ & $\text{26.73}_{\hspace{0.05cm}0.29}^{*}$ & $\text{28.54}_{\hspace{0.05cm}0.46}^{*}$ & $\text{34.83}_{\hspace{0.05cm}0.20}^{*}$ & $\text{42.68}_{\hspace{0.05cm}0.13}^{*}$ & $\text{37.07}^{*}$ / $\text{40.76}^{*}$ \\

\multicolumn{12}{l}{\LogoMiddle\ \textsc{LM-Lexicon-MoE (4$\times$8B)}} \\
~ \hspace{-0.2cm}$\hookrightarrow \text{+}$ \textit{Zero-shot} & $\underline{40.09}_{\hspace{0.05cm}0.12}^{*}$ & $\underline{40.51}_{\hspace{0.05cm}0.28}^{*}$ & $\text{23.35}_{\hspace{0.05cm}0.25}$ & $\text{32.94}_{\hspace{0.05cm}0.49}^{*}$ & $\underline{60.31}_{\hspace{0.05cm}0.55}^{*}$ & $\text{55.52}_{\hspace{0.05cm}0.33}$ & $\textbf{31.26}_{\hspace{0.05cm}0.85}^{*}$ & $\textbf{33.81}_{\hspace{0.05cm}2.26}^{*}$ & $\underline{45.69}_{\hspace{0.05cm}1.25}^{*}$ & $\underline{46.07}_{\hspace{0.05cm}1.06}^{*}$ & $\underline{40.14}^{*}$ / $\underline{41.77}^{*}$ \\
 ~ \hspace{-0.2cm}$\hookrightarrow \text{+}$ \textit{BoN-Oracle}$^{\dagger}$ & \textcolor{Gray!75}{$\text{47.39}_{\hspace{0.05cm}0.16}$} & \textcolor{Gray!75}{$\text{40.31}_{\hspace{0.05cm}0.23}$} & \textcolor{Gray!75}{$\text{30.87}_{\hspace{0.05cm}0.24}$} & \textcolor{Gray!75}{$\text{43.24}_{\hspace{0.05cm}0.25}$} & \textcolor{Gray!75}{$\text{51.62}_{\hspace{0.05cm}1.14}$} & \textcolor{Gray!75}{$\text{61.88}_{\hspace{0.05cm}0.30}$} & \textcolor{Gray!75}{$\text{35.23}_{\hspace{0.05cm}0.42}$} & \textcolor{Gray!75}{$\text{35.69}_{\hspace{0.05cm}0.26}$} & \textcolor{Gray!75}{$\text{54.84}_{\hspace{0.05cm}0.12}$} & \textcolor{Gray!75}{$\text{50.50}_{\hspace{0.05cm}0.11}$} & \textcolor{Gray!75}{$\text{43.99}$} \textcolor{Gray!75}{/} \textcolor{Gray!75}{$\text{46.32}$} \\
 ~ \hspace{-0.2cm}$\hookrightarrow \text{+}$ \textit{BoN-ORM} & $\textbf{40.33}_{\hspace{0.05cm}0.18}^{*}$ & $40.69_{\hspace{0.05cm}0.26}^{*}$ & $\text{24.18}_{\hspace{0.05cm}0.37}$ & $\text{33.79}_{\hspace{0.05cm}0.64}^{*}$ & $\textbf{60.88}_{\hspace{0.05cm}0.55}^{*}$ & $57.66_{\hspace{0.05cm}0.73}$ & $\underline{\text{31.08}}_{\hspace{0.05cm}0.17}^{*}$ & $\underline{\text{33.26}}_{\hspace{0.05cm}0.22}^{*}$ & $\textbf{45.86}_{\hspace{0.05cm}0.38}^{*}$ & $\textbf{46.38}_{\hspace{0.05cm}0.26}^{*}$ & $\textbf{40.46}^{*}$ / $\textbf{42.35}^{*}$ \\

\bottomrule
\end{tabular}}
\end{center}

\caption{Main results on five benchmarks{\protect\footnotemark}. We highlight the \textbf{highest scores} among \ourmethod and compared methods; * denotes the significance test, where $p < 0.005$ between our method and Rerank-T5 (prior SoTA). $\clubsuit$ denotes that we reproduce the in-distribution results with supervised training, and $\dagger$ indicates that the lines of results are not directly comparable with other settings. All \textit{*-ICL} settings employ the best setting with a 32-shot in practice.}
\label{table:main_results}
\end{table*}
\footnotetext{We develop ad-hoc heuristic parser for proprietary models \& \ourmethod to extract our focused part of the generation.}
\paragraph{Competitive Performance of \LogoSmall\ \ourmethod.} Table \ref{table:main_results} presents the performance comparisons among baselines and existing SoTA methods for DM, including \textsc{LM-Lexicon-Dense} models (trained on four real-world datasets) and \textsc{LM-Lexicon-MoE}, the proposed MoE model. \LogoSmall\ \ourmethod outperforms strong supervised methods and frontier models with a distinct advantage. Specifically, (1) \LogoSmall\ \ourmethod obtains nearly $10\%$ extra BLEU and ROUGE improvements on 3D-EX over the prior SoTA. (2) It performs exceptionally on smaller datasets as well, for example, \LogoSmall\ \ourmethod achieves the highest scores ($\{31.26\%, 33.81\%\}$ on \{BLEU, ROUGE\}) among all compared methods on Urban dataset, indicating the efficacy of our method to model rare word senses and usages. (3) 
The comparison between the many-shot learning of best perfomant frontier LMs and \LogoSmall\ \ourmethod demonstrates that our method surpasses significantly larger dense models, by \{$23.44\%$, $9.14\%$\} on \{Wiki, WordNet\} in BLEU for instance. (4) It is also observed that the Oxford dataset has lower performance with our method. A possible reason is that a short term and relatively long context in Oxford makes it harder for the model to predict accurate definitions. Furthermore, compared to other benchmarks, the Oxford dataset exhibits a significantly high term overlap rate of around 80\% along with a near-zero term-definition overlap rate. This stark contrast underscores the strong polysemy inherent in Oxford's terms. Consequently, models trained on Oxford struggle to generalize effectively when encountering previously seen terms used in different contexts. Overall, \LogoSmall\ \ourmethod shows a clear advantage that confirms the effectiveness of introduced semantic expert specialization and semantic-focused sparsifying upcycling into \LogoSmall\ \ourmethod.

\paragraph{Human Evaluation.} The human evaluation was conducted using a random subset of 300 samples from the 3D-EX, comparing definitions generated by our model (\textsc{LM-Lexicon-MoE}) and the baselines (\textsc{LM-Lexicon-Dense} and three proprietary models). We focus on comparing with proprietary models as they represent the current state-of-the-art in practical deployment and are the primary competitors in real-world lexicon construction scenarios. To obtain a fine-grained understanding of model-specific characteristics, we further propose five criteria: (1) \textit{accuracy} measures how correctly the definition captures the core semantic meaning of the word; (2) \textit{clarity} evaluates the definition's comprehensibility and transparency in conveying meaning, focusing on how easily readers can understand the concept; (3) \textit{conciseness} assesses whether the definition achieves optimal length without redundancy or omission; (4) \textit{context appropriateness} measures how well the definition reflects associated contexts, situations, and pragmatic constraints of the words; (5) \textit{grammar and fluency} evaluates the grammatical correctness and naturalness of the definition. We employ three graduate students majoring in linguistics and lexicography, who were instructed to assess each of the above criteria on a 5-point scale, where 1 indicates the poorest quality and 5 represents the highest quality (Figure \ref{fig:human_eval_guideline}). The model names were kept anonymous from human evaluators to avoid possible bias, whereas the reference definitions remained accessible to them. 

Figure \ref{fig:semantic_scaling_and_human_eval} (right) presents the human evaluation results across five criteria, showing the average scores for each model\footnote{Details on annotators' agreement can be found in \S\ref{sec:annotators' agreement}.}. \textsc{LM-Lexicon-MoE} consistently outperforms other models in most dimensions, with particularly strong performance of accuracy (4.6). While all models demonstrate competent performance with scores above 3.8, \textsc{LM-Lexicon-MoE} shows notable advantages in capturing contextual nuances and maintaining clarity and conciseness in definitions. The proprietary models perform similarly well but show slightly lower scores in terms of context appropriateness and conciseness than other criteria. We provide a detailed analysis of a representative example ``\textit{coon}'' in Appendix \ref{sec:representative_case}.
\begin{figure}[t]
    \includegraphics[width=\linewidth]{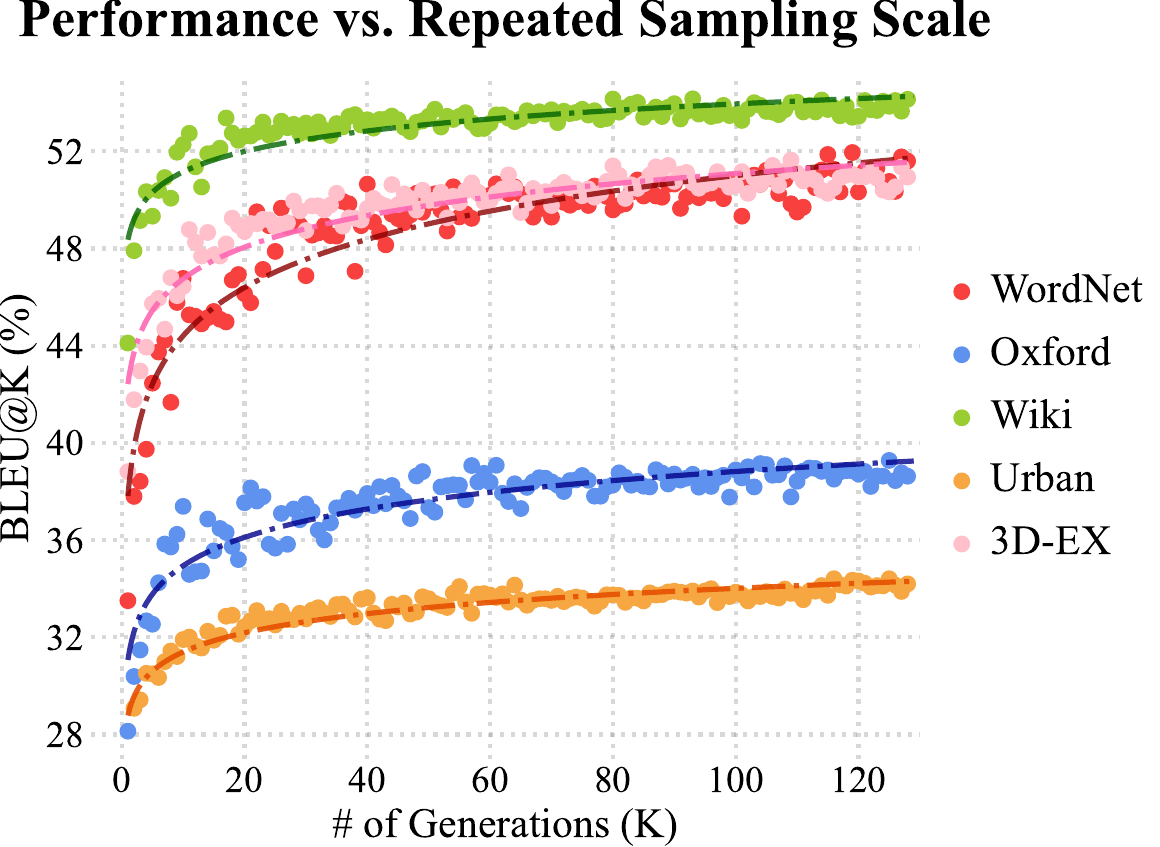}
  \caption{\textit{Best-of-N} repeated sampling results (\textsc{Bleu}) on five benchmarks evaluated by \textbf{oracle verifier}.}
  \label{fig:repeated_sampling_performance}
\end{figure}

\subsection{Ablation Study and Extra Investigation}
\label{subsec:ablation_study}
In this section, we further conduct an in-depth analysis of \LogoSmall\ \ourmethod, regarding: (1) data partition method, (2) routing policy, and (3) number of experts. In addition, we explore the impact of test-time scaling. Finally, we examine the scaling effect of ICL for proprietary LLMs.
\begin{figure*}[ht]
    \centering
    \begin{minipage}{0.44\linewidth}
        \centering
        \includegraphics[width=\linewidth]{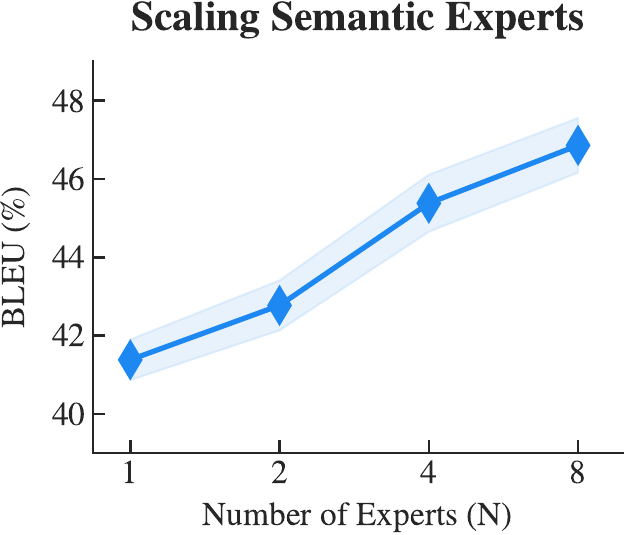}
    \end{minipage}
    \hfill
    \begin{minipage}{0.535\linewidth}
        \centering
        \includegraphics[width=\linewidth]{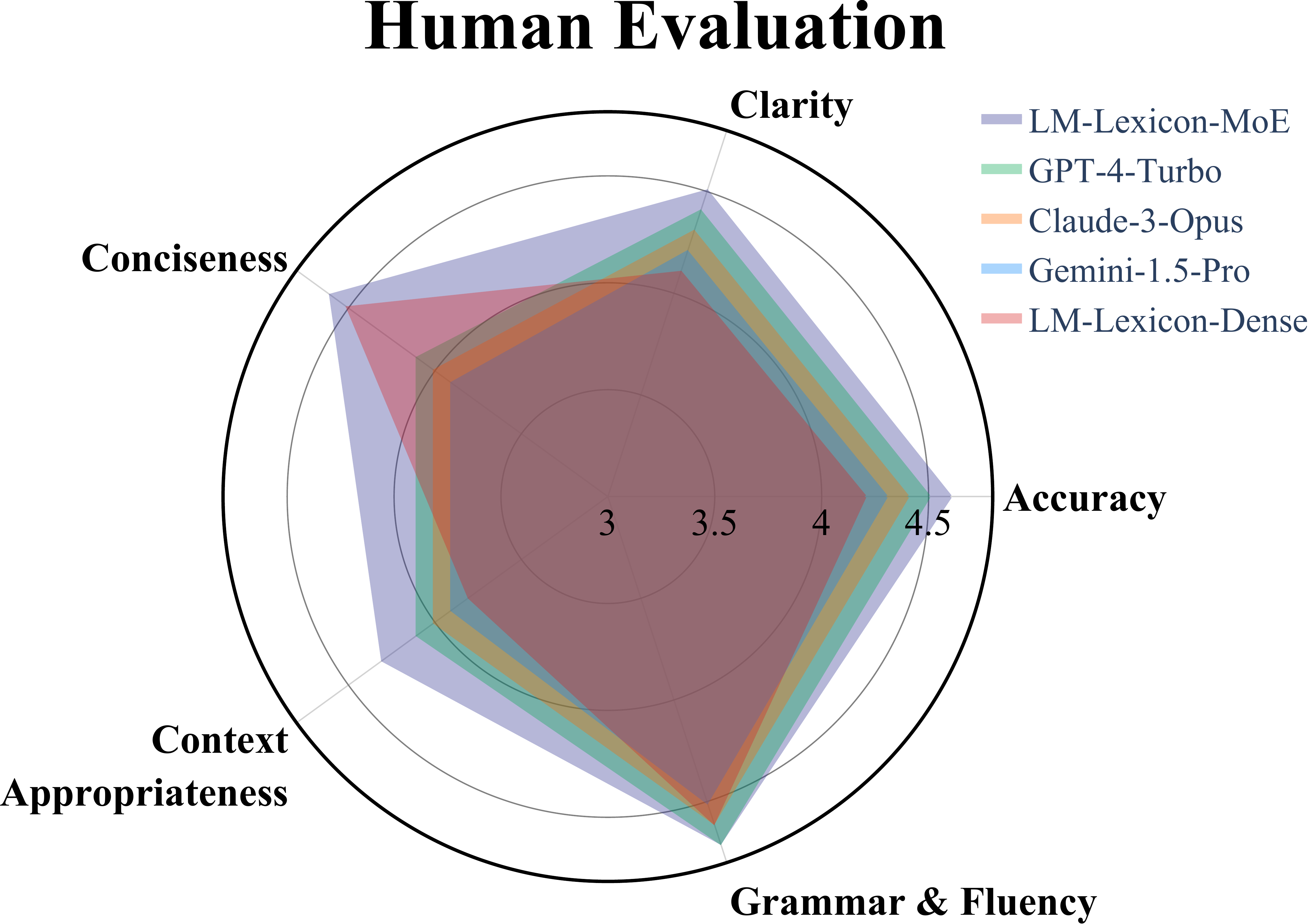}
    \end{minipage}
    \caption{Scaling performance gains and human evaluation results. The left figure: Scaling test performance on 3D-EX, with varying number of experts. The right figure: Human evaluation results across five criteria.}
    \label{fig:semantic_scaling_and_human_eval}
\end{figure*}
\paragraph{Ablation on Different Data Partition Designs.}
\label{para:cluster-policy-ablation}
Since \LogoSmall\ \ourmethod integrates the knowledge acquired by experts from various data partitions, our first focus is on the impact of data partition methods. To this end, we considered three settings: (1) no split; (2) random split; and (3) lexical split. For random split, we follow \citet{li2022branch} to slice the data into four balanced subsets and specialise an expert for each of them. For lexical split, we perform partition by TF-IDF \cite{sparck1972statistical}.

As shown in Table \ref{tab:ablation-data-partition}, we observed that the original setting with semantic embedding clustering outperforms lexical-based partition with about $+7\%$ gains in \textsc{Bleu} and $+1\%$ gains in \textsc{Rouge} on 3D-EX. The results imply that learning from semantic-targeted data clusters may help capture more precise senses and use more appropriate words to compose definitions. Lastly, it enables \LogoSmall\ \ourmethod to develop more robust experts for various domains.

\begin{table}[ht]
    \centering
    \small
    \vspace{-1em}
    {\begin{tabular}{l|ccl}
        \multicolumn{2}{c}{} \\
        \toprule
        \textbf{Model} & \multicolumn{1}{c}{\textbf{BLEU}} & \multicolumn{1}{c}{\textbf{ROUGE}} & \multicolumn{1}{c}{\textbf{p-value}} \\
        \midrule
        \LogoSmall\ { \small\ourmethod} & 45.69${\scriptscriptstyle\pm}$\tiny{0.3} & 46.07${\scriptscriptstyle\pm}$\tiny{0.1} & $-$ \\
        \midrule
        + w/ no split & 35.13${\scriptscriptstyle\pm}$\tiny{0.2} & 43.46${\scriptscriptstyle\pm}$\tiny{0.3} & $2.9e^{-5}$ \\
        + w/ random split & 36.24${\scriptscriptstyle\pm}$\tiny{1.4} & 43.58${\scriptscriptstyle\pm}$\tiny{0.8} & $1.6e^{-5}$ \\
        + w/ lexical split & 38.13${\scriptscriptstyle\pm}$\tiny{0.5} & 44.12${\scriptscriptstyle\pm}$\tiny{0.6} & $1.3e^{-4}$ \\
        \bottomrule
    \end{tabular}}
    \caption{Ablation on data partition method.}
    \label{tab:ablation-data-partition}
\end{table}

\paragraph{Comparison among Routing Policies.}
\label{para:routing-strategy-ablation}
Other than domain-level routing used in \LogoSmall\ \ourmethod as default, we experiment on (1) top-1 token-level; (2) top-2 token-level; and (3) sequence-level routing. For token-level routing, we follow the implementation of \citet{fedus2022switch} and \citet{jiang2024mixtral}. For sequence-level routing, we follow \citet{pham-etal-2023-task}.
\begin{table}[ht]
    \centering
    \small
    {\begin{tabular}{l|ccl}
        \multicolumn{2}{c}{} \\
        \toprule
        \textbf{Model} & \multicolumn{1}{c}{\textbf{BLEU}} & \multicolumn{1}{c}{\textbf{ROUGE}} & \multicolumn{1}{c}{\textbf{p-value}} \\
        \midrule
        \LogoSmall\ { \small\ourmethod} & 45.69${\scriptscriptstyle\pm}$\tiny{0.3} & 46.07${\scriptscriptstyle\pm}$\tiny{0.1} & $-$ \\
        \midrule
        + w/ top-1 token-level & 43.12${\scriptscriptstyle\pm}$\tiny{0.4} & 43.79${\scriptscriptstyle\pm}$\tiny{0.5} & $1.9e^{-3}$ \\
        + w/ top-2 token-level & 45.38${\scriptscriptstyle\pm}$\tiny{0.2} & 45.21${\scriptscriptstyle\pm}$\tiny{0.1} & $8.6e^{-1}$ \\
        + w/ sequence-level & 44.47${\scriptscriptstyle\pm}$\tiny{0.2} & 44.82${\scriptscriptstyle\pm}$\tiny{0.3} & $2.7e^{-3}$ \\
        \bottomrule
    \end{tabular}}
    \caption{Ablation on different routing policies.}
    \label{tab:ablation-routing-policy}
\end{table}

Table \ref{tab:ablation-routing-policy} presents that the domain-level routing (\LogoSmall\ \ourmethod) is the most effective, even surpassing one of the popular scheme, the top-2 token-level routing, indicating that semantic routing via specified domain cluster is more beneficial for semantic-intensive tasks.
\paragraph{Different Number of Semantic Experts.}
\label{para:expert-numbers-ablation}
Except for the above four-experts \textsc{LM-Lexicon-MoE}, to investigate the impact of the number of semantic experts, we compare varied number of semantic experts ($N = 1, 2, 4, 8$). Notably, when $N=1$, \LogoSmall\ \ourmethod collapses back to a dense model and expands to a sparse model with $N > 1$ experts.

As shown in Figure \ref{fig:semantic_scaling_and_human_eval} (left), we find that across all settings of $N$, the performance of our method consistently increases and outperforms the others, which are composed of fewer experts. For example, the model of $N=1$ returns $41.38\%$ while $N=8$ yields $46.86\%$ in \textsc{Bleu}. This tendency implies the scalability of our method, using more semantic experts. This trend can be extended by integrating more fine-grained semantic experts \cite{dai-etal-2024-deepseekmoe}, but we leave this direction for future work.

\paragraph{Impact of Test-time Scaling.}
In light of \citet{NEURIPS2020_1f89885d} and \citet{cobbe2021training}, we are curious on how to boost performance further via test-time scaling, notably ground truth-based (\textit{i.e.}, Oracle) verifier and Best-of-N (BoN) sampling with an outcome reward model (ORM). For oracle verifier, it uses reference as verification to provide binary feedbacks. For an ORM, it employs scalar feedback to select the optimal generation from candidates.

As depicted in Table \ref{table:main_results} (BoN-ORM), interestingly, the oracle verifier is able to boost task performance (avg. $\Delta\textsc{Bleu} > 2\%$) for \textsc{LM-Lexicon-Dense}. However, it exhibits more limitations for \textsc{LM-Lexicon-MoE}; we speculate it is due to the diversity diminishment of models, as illustrated in \citet{brown2024large}. Intuitively, optimal results are achieved with oracle verifier (Fig. \ref{fig:repeated_sampling_performance}) through repeated sampling with 128 completions per test sample. Intergating with the ORM or Oracle verifier, \LogoSmall\ \ourmethod's generation quality shows consistent improvements across five benchmarks with the increase in the number of generations. This outcome aligns with the findings on math reasoning tasks \cite{cobbe2021training,brown2024large}.

\section{Conclusion}\label{sec:conclusion}
In this paper, we present \LogoSmall\ \ourmethod, an approach that combines domain experts upcycling with a sparse MoE model, which can generate appropriate definitions of terms in various domains and genres. We show that it significantly outperforms frontier LLMs and strong supervised baselines. We hope \LogoSmall\ \ourmethod could be extended to more domains and other semantic-intensive tasks in the future.
\section*{Limitations}\label{sec:limitation}

\paragraph{Extrapolation to More Tasks.}
While we believe our observations and conclusions are comprehensive within our experimental settings, our work only focus on the task of definition modeling in English in this work. Future work could benefit from our findings in extending to other domains and similar tasks in semantic-intensive scenarios.

\paragraph{Training Efficienty and Cost.} Our method performs supervised fine-tuning of $N\times\mathcal{M}$ expert LMs that are initialized from a seed model. The training process can be thoroughly offline and asynchronous; however, it still needs an essential and sufficient computation budget to some extent. We encourage people to further explore parameter-efficient training methods based on \ourmethod.

\paragraph{Desiderata for Stronger Verifiers.} Our results from Section \S\ref{subsec:ablation_study} highlight the significance of improving sample verification methods tailored for definition modeling, and even more general language generation, which are currently unavailable or highly limited. Most existing verification methods have been developed only to solve easily verifiable reasoning tasks, such as mathematical \cite{li-etal-2025-reflectevo}, software engineering \cite{yang2024sweagent}, and logical reasoning problems \cite{liu2025rulereasoner}. We believe that equipping models with the ability to assess their own generations will allow test-time compute methods to be scaled further.
\section*{Ethics Statement}
\label{sec:ethics-statement}
This research was conducted with careful consideration of ethical implications. All data used in this study was collected from public sources with appropriate permissions. We have taken measures to ensure privacy protection and prevent misuse of our model. The computational resources were used responsibly, and we have documented all potential biases and limitations. Our annotation process followed fair labor practices with appropriate compensation for annotators.
\section*{Acknowledgement}
\label{sec:acknowledgement}
We are deeply grateful to all the reviewers for their valuable feedback and thoughtful efforts in helping us improve this manuscript. We would also like to thank Ziang Wu for his contributions to the early exploration and discussions that shaped this work, and Ivan Fung for his support of the computational resources that made this project possible.

\bibliography{custom}

\begin{thebibliography}{62}
\providecommand{\natexlab}[1]{#1}

\bibitem[{Achiam et~al.(2023)Achiam, Adler, Agarwal, Ahmad, Akkaya, Aleman, Almeida, Altenschmidt, Altman, Anadkat et~al.}]{achiam2023gpt}
Josh Achiam, Steven Adler, Sandhini Agarwal, Lama Ahmad, Ilge Akkaya, Florencia~Leoni Aleman, Diogo Almeida, Janko Altenschmidt, Sam Altman, Shyamal Anadkat, and 1 others. 2023.
\newblock Gpt-4 technical report.
\newblock \emph{arXiv preprint arXiv:2303.08774}.

\bibitem[{Ahlswede(1985)}]{ahlswede-1985-tool}
Thomas~E. Ahlswede. 1985.
\newblock \href {https://doi.org/10.3115/981210.981243} {A tool kit for lexicon building}.
\newblock In \emph{23rd Annual Meeting of the Association for Computational Linguistics}, pages 268--276, Chicago, Illinois, USA. Association for Computational Linguistics.

\bibitem[{Almeman et~al.(2023)Almeman, Sheikhi, and Espinosa~Anke}]{almeman-etal-2023-3d}
Fatemah Almeman, Hadi Sheikhi, and Luis Espinosa~Anke. 2023.
\newblock \href {https://aclanthology.org/2023.ranlp-1.8} {3{D}-{EX}: A unified dataset of definitions and dictionary examples}.
\newblock In \emph{Proceedings of the 14th International Conference on Recent Advances in Natural Language Processing}, pages 69--79, Varna, Bulgaria. INCOMA Ltd., Shoumen, Bulgaria.

\bibitem[{Almeman et~al.(2024)Almeman, Schockaert, and Espinosa~Anke}]{almeman-etal-2024-wordnet}
Fatemah~Yousef Almeman, Steven Schockaert, and Luis Espinosa~Anke. 2024.
\newblock \href {https://aclanthology.org/2024.lrec-main.1538} {{W}ord{N}et under scrutiny: Dictionary examples in the era of large language models}.
\newblock In \emph{Proceedings of the 2024 Joint International Conference on Computational Linguistics, Language Resources and Evaluation (LREC-COLING 2024)}, pages 17683--17695, Torino, Italia. ELRA and ICCL.

\bibitem[{Anthropic(2024)}]{anthropic2024claude}
AI~Anthropic. 2024.
\newblock The claude 3 model family: Opus, sonnet, haiku.
\newblock \emph{Claude-3 Model Card}.

\bibitem[{Bai et~al.(2025)Bai, Tong, Liu, Jia, and Zheng}]{bai-etal-2025-understanding}
Jun Bai, Minghao Tong, Yang Liu, Zixia Jia, and Zilong Zheng. 2025.
\newblock \href {https://doi.org/10.18653/v1/2025.emnlp-main.1114} {Understanding and leveraging the expert specialization of context faithfulness in mixture-of-experts {LLM}s}.
\newblock In \emph{Proceedings of the 2025 Conference on Empirical Methods in Natural Language Processing}, pages 21927--21942, Suzhou, China. Association for Computational Linguistics.

\bibitem[{Bricken et~al.(2023)Bricken, Templeton, Batson, Chen, Jermyn, Conerly, Turner, Anil, Denison, Askell, Lasenby, Wu, Kravec, Schiefer, Maxwell, Joseph, Hatfield-Dodds, Tamkin, Nguyen, McLean, Burke, Hume, Carter, Henighan, and Olah}]{bricken2023monosemanticity}
Trenton Bricken, Adly Templeton, Joshua Batson, Brian Chen, Adam Jermyn, Tom Conerly, Nick Turner, Cem Anil, Carson Denison, Amanda Askell, Robert Lasenby, Yifan Wu, Shauna Kravec, Nicholas Schiefer, Tim Maxwell, Nicholas Joseph, Zac Hatfield-Dodds, Alex Tamkin, Karina Nguyen, and 6 others. 2023.
\newblock Towards monosemanticity: Decomposing language models with dictionary learning.
\newblock \emph{Transformer Circuits Thread}.
\newblock Https://transformer-circuits.pub/2023/monosemantic-features/index.html.

\bibitem[{Brown et~al.(2024)Brown, Juravsky, Ehrlich, Clark, Le, R{\'e}, and Mirhoseini}]{brown2024large}
Bradley Brown, Jordan Juravsky, Ryan Ehrlich, Ronald Clark, Quoc~V Le, Christopher R{\'e}, and Azalia Mirhoseini. 2024.
\newblock Large language monkeys: Scaling inference compute with repeated sampling.
\newblock \emph{arXiv preprint arXiv:2407.21787}.

\bibitem[{Cobbe et~al.(2021)Cobbe, Kosaraju, Bavarian, Chen, Jun, Kaiser, Plappert, Tworek, Hilton, Nakano et~al.}]{cobbe2021training}
Karl Cobbe, Vineet Kosaraju, Mohammad Bavarian, Mark Chen, Heewoo Jun, Lukasz Kaiser, Matthias Plappert, Jerry Tworek, Jacob Hilton, Reiichiro Nakano, and 1 others. 2021.
\newblock Training verifiers to solve math word problems.
\newblock \emph{arXiv preprint arXiv:2110.14168}.

\bibitem[{Dai et~al.(2024)Dai, Deng, Zhao, Xu, Gao, Chen, Li, Zeng, Yu, Wu, Xie, Li, Huang, Luo, Ruan, Sui, and Liang}]{dai-etal-2024-deepseekmoe}
Damai Dai, Chengqi Deng, Chenggang Zhao, R.x. Xu, Huazuo Gao, Deli Chen, Jiashi Li, Wangding Zeng, Xingkai Yu, Y.~Wu, Zhenda Xie, Y.k. Li, Panpan Huang, Fuli Luo, Chong Ruan, Zhifang Sui, and Wenfeng Liang. 2024.
\newblock \href {https://doi.org/10.18653/v1/2024.acl-long.70} {{D}eep{S}eek{M}o{E}: Towards ultimate expert specialization in mixture-of-experts language models}.
\newblock In \emph{Proceedings of the 62nd Annual Meeting of the Association for Computational Linguistics (Volume 1: Long Papers)}, pages 1280--1297, Bangkok, Thailand. Association for Computational Linguistics.

\bibitem[{Dubey et~al.(2024)Dubey, Jauhri, Pandey, Kadian, Al-Dahle, Letman, Mathur, Schelten, Yang, Fan et~al.}]{dubey2024llama}
Abhimanyu Dubey, Abhinav Jauhri, Abhinav Pandey, Abhishek Kadian, Ahmad Al-Dahle, Aiesha Letman, Akhil Mathur, Alan Schelten, Amy Yang, Angela Fan, and 1 others. 2024.
\newblock The llama 3 herd of models.
\newblock \emph{arXiv preprint arXiv:2407.21783}.

\bibitem[{Elhage et~al.(2022)Elhage, Hume, Olsson, Schiefer, Henighan, Kravec, Hatfield-Dodds, Lasenby, Drain, Chen, Grosse, McCandlish, Kaplan, Amodei, Wattenberg, and Olah}]{elhage2022superposition}
Nelson Elhage, Tristan Hume, Catherine Olsson, Nicholas Schiefer, Tom Henighan, Shauna Kravec, Zac Hatfield-Dodds, Robert Lasenby, Dawn Drain, Carol Chen, Roger Grosse, Sam McCandlish, Jared Kaplan, Dario Amodei, Martin Wattenberg, and Christopher Olah. 2022.
\newblock \href {https://transformer-circuits.pub/2022/toy_model/index.html} {Toy models of superposition}.
\newblock \emph{Transformer Circuits Thread}.

\bibitem[{Fedus et~al.(2022)Fedus, Zoph, and Shazeer}]{fedus2022switch}
William Fedus, Barret Zoph, and Noam Shazeer. 2022.
\newblock Switch transformers: Scaling to trillion parameter models with simple and efficient sparsity.
\newblock \emph{Journal of Machine Learning Research}, 23(120):1--39.

\bibitem[{Fleiss(1971)}]{Fleiss_1971}
J.~L. Fleiss. 1971.
\newblock \href {https://doi.org/10.1037/h0031619} {Measuring nominal scale agreement among many raters}.
\newblock \emph{Psychological Bulletin}, 76(5):378–382.

\bibitem[{Gadetsky et~al.(2018)Gadetsky, Yakubovskiy, and Vetrov}]{gadetsky2018conditional}
A~Gadetsky, I~Yakubovskiy, and D~Vetrov. 2018.
\newblock Conditional generators of words definitions.
\newblock In \emph{ACL 2018-56th Annual Meeting of the Association for Computational Linguistics, Proceedings of the Conference (Long Papers)}, pages 266--271.

\bibitem[{Giulianelli et~al.(2023)Giulianelli, Luden, Fernandez, and Kutuzov}]{giulianelli-etal-2023-interpretable}
Mario Giulianelli, Iris Luden, Raquel Fernandez, and Andrey Kutuzov. 2023.
\newblock \href {https://doi.org/10.18653/v1/2023.acl-long.176} {Interpretable word sense representations via definition generation: The case of semantic change analysis}.
\newblock In \emph{Proceedings of the 61st Annual Meeting of the Association for Computational Linguistics (Volume 1: Long Papers)}, pages 3130--3148, Toronto, Canada. Association for Computational Linguistics.

\bibitem[{Gururangan et~al.(2023)Gururangan, Li, Lewis, Shi, Althoff, Smith, and Zettlemoyer}]{gururangan2023scaling}
Suchin Gururangan, Margaret Li, Mike Lewis, Weijia Shi, Tim Althoff, Noah~A Smith, and Luke Zettlemoyer. 2023.
\newblock Scaling expert language models with unsupervised domain discovery.
\newblock \emph{arXiv preprint arXiv:2303.14177}.

\bibitem[{Hogeweg and Vicente(2020)}]{hogeweg2020nature}
Lotte Hogeweg and Agustin Vicente. 2020.
\newblock On the nature of the lexicon: The status of rich lexical meanings.
\newblock \emph{Journal of Linguistics}, 56(4):865--891.

\bibitem[{Huang et~al.(2021)Huang, Kajiwara, and Arase}]{huang-etal-2021-definition}
Han Huang, Tomoyuki Kajiwara, and Yuki Arase. 2021.
\newblock \href {https://doi.org/10.18653/v1/2021.emnlp-main.194} {Definition modelling for appropriate specificity}.
\newblock In \emph{Proceedings of the 2021 Conference on Empirical Methods in Natural Language Processing}, pages 2499--2509, Online and Punta Cana, Dominican Republic. Association for Computational Linguistics.

\bibitem[{Ishiwatari et~al.(2019)Ishiwatari, Hayashi, Yoshinaga, Neubig, Sato, Toyoda, and Kitsuregawa}]{ishiwatari-etal-2019-learning}
Shonosuke Ishiwatari, Hiroaki Hayashi, Naoki Yoshinaga, Graham Neubig, Shoetsu Sato, Masashi Toyoda, and Masaru Kitsuregawa. 2019.
\newblock \href {https://doi.org/10.18653/v1/N19-1350} {Learning to describe unknown phrases with local and global contexts}.
\newblock In \emph{Proceedings of the 2019 Conference of the North {A}merican Chapter of the Association for Computational Linguistics: Human Language Technologies, Volume 1 (Long and Short Papers)}, pages 3467--3476, Minneapolis, Minnesota. Association for Computational Linguistics.

\bibitem[{Jhirad et~al.(2023)Jhirad, Marrese-Taylor, and Matsuo}]{jhirad2023evaluating}
James Jhirad, Edison Marrese-Taylor, and Yutaka Matsuo. 2023.
\newblock Evaluating large language models’ understanding of financial terminology via definition modeling.
\newblock In \emph{Proceedings of the 13th International Joint Conference on Natural Language Processing and the 3rd Conference of the Asia-Pacific Chapter of the Association for Computational Linguistics: Student Research Workshop}, pages 93--100.

\bibitem[{Jiang et~al.(2024)Jiang, Sablayrolles, Roux, Mensch, Savary, Bamford, Chaplot, Casas, Hanna, Bressand et~al.}]{jiang2024mixtral}
Albert~Q Jiang, Alexandre Sablayrolles, Antoine Roux, Arthur Mensch, Blanche Savary, Chris Bamford, Devendra~Singh Chaplot, Diego de~las Casas, Emma~Bou Hanna, Florian Bressand, and 1 others. 2024.
\newblock Mixtral of experts.
\newblock \emph{arXiv preprint arXiv:2401.04088}.

\bibitem[{Kong et~al.(2022)Kong, Chen, Zhang, Yang, and Yang}]{kong-etal-2022-multitasking}
Cunliang Kong, Yun Chen, Hengyuan Zhang, Liner Yang, and Erhong Yang. 2022.
\newblock \href {https://doi.org/10.18653/v1/2022.acl-long.409} {Multitasking framework for unsupervised simple definition generation}.
\newblock In \emph{Proceedings of the 60th Annual Meeting of the Association for Computational Linguistics (Volume 1: Long Papers)}, pages 5934--5943, Dublin, Ireland. Association for Computational Linguistics.

\bibitem[{Lavie and Agarwal(2007)}]{lavie-agarwal-2007-meteor}
Alon Lavie and Abhaya Agarwal. 2007.
\newblock \href {https://aclanthology.org/W07-0734} {{METEOR}: An automatic metric for {MT} evaluation with high levels of correlation with human judgments}.
\newblock In \emph{Proceedings of the Second Workshop on Statistical Machine Translation}, pages 228--231, Prague, Czech Republic. Association for Computational Linguistics.

\bibitem[{Lee et~al.(2025)Lee, Roy, Xu, Raiman, Shoeybi, Catanzaro, and Ping}]{lee2025nvembedimprovedtechniquestraining}
Chankyu Lee, Rajarshi Roy, Mengyao Xu, Jonathan Raiman, Mohammad Shoeybi, Bryan Catanzaro, and Wei Ping. 2025.
\newblock \href {https://arxiv.org/abs/2405.17428} {Nv-embed: Improved techniques for training llms as generalist embedding models}.
\newblock \emph{Preprint}, arXiv:2405.17428.

\bibitem[{{Leeroo-AI}(2024)}]{mergoo2024}
{Leeroo-AI}. 2024.
\newblock Mergoo: A library for easily merging multiple llm experts, and efficiently train the merged llm.
\newblock \url{https://github.com/Leeroo-AI/mergoo}.
\newblock Accessed: 2024-07-23.

\bibitem[{Li et~al.(2025)Li, Dong, Liu, Yang, Wang, Wang, Zhu, Jia, and Zheng}]{li-etal-2025-reflectevo}
Jiaqi Li, Xinyi Dong, Yang Liu, Zhizhuo Yang, Quansen Wang, Xiaobo Wang, Song-Chun Zhu, Zixia Jia, and Zilong Zheng. 2025.
\newblock \href {https://doi.org/10.18653/v1/2025.findings-acl.871} {{R}eflect{E}vo: Improving meta introspection of small {LLM}s by learning self-reflection}.
\newblock In \emph{Findings of the Association for Computational Linguistics: ACL 2025}, pages 16948--16966, Vienna, Austria. Association for Computational Linguistics.

\bibitem[{Li et~al.(2022)Li, Gururangan, Dettmers, Lewis, Althoff, Smith, and Zettlemoyer}]{li2022branch}
Margaret Li, Suchin Gururangan, Tim Dettmers, Mike Lewis, Tim Althoff, Noah~A Smith, and Luke Zettlemoyer. 2022.
\newblock Branch-train-merge: Embarrassingly parallel training of expert language models.
\newblock \emph{arXiv preprint arXiv:2208.03306}.

\bibitem[{Lin(2004)}]{lin-2004-rouge}
Chin-Yew Lin. 2004.
\newblock \href {https://aclanthology.org/W04-1013} {{ROUGE}: A package for automatic evaluation of summaries}.
\newblock In \emph{Text Summarization Branches Out}, pages 74--81, Barcelona, Spain. Association for Computational Linguistics.

\bibitem[{Liu et~al.(2025)Liu, Li, and Zheng}]{liu2025rulereasoner}
Yang Liu, Jiaqi Li, and Zilong Zheng. 2025.
\newblock Rulereasoner: Reinforced rule-based reasoning via domain-aware dynamic sampling.
\newblock \emph{arXiv preprint arXiv:2506.08672}.

\bibitem[{Loshchilov and Hutter(2018)}]{loshchilov2018decoupled}
Ilya Loshchilov and Frank Hutter. 2018.
\newblock Decoupled weight decay regularization.
\newblock In \emph{International Conference on Learning Representations}.

\bibitem[{Ma et~al.(2024)Ma, Huang, Xie, Li, Zettlemoyer, Chang, Yih, and Xu}]{ma2024mode}
Jiawei Ma, Po-Yao Huang, Saining Xie, Shang-Wen Li, Luke Zettlemoyer, Shih-Fu Chang, Wen-Tau Yih, and Hu~Xu. 2024.
\newblock Mode: Clip data experts via clustering.
\newblock In \emph{Proceedings of the IEEE/CVF conference on computer vision and pattern recognition}, pages 26354--26363.

\bibitem[{Malinen and Fr{\"a}nti(2014)}]{balancedkmeans}
Mikko~I. Malinen and Pasi Fr{\"a}nti. 2014.
\newblock Balanced k-means for clustering.
\newblock In \emph{Structural, Syntactic, and Statistical Pattern Recognition}, pages 32--41, Berlin, Heidelberg. Springer Berlin Heidelberg.

\bibitem[{Ni and Wang(2017)}]{ni2017learning}
Ke~Ni and William~Yang Wang. 2017.
\newblock Learning to explain non-standard english words and phrases.
\newblock In \emph{Proceedings of the Eighth International Joint Conference on Natural Language Processing (Volume 2: Short Papers)}, pages 413--417.

\bibitem[{Noraset et~al.(2017)Noraset, Liang, Birnbaum, and Downey}]{noraset2017definition}
Thanapon Noraset, Chen Liang, Larry Birnbaum, and Doug Downey. 2017.
\newblock Definition modeling: Learning to define word embeddings in natural language.
\newblock In \emph{Proceedings of the AAAI Conference on Artificial Intelligence}, volume~31.

\bibitem[{Papineni et~al.(2002)Papineni, Roukos, Ward, and Zhu}]{papineni-etal-2002-bleu}
Kishore Papineni, Salim Roukos, Todd Ward, and Wei-Jing Zhu. 2002.
\newblock \href {https://doi.org/10.3115/1073083.1073135} {{B}leu: a method for automatic evaluation of machine translation}.
\newblock In \emph{Proceedings of the 40th Annual Meeting of the Association for Computational Linguistics}, pages 311--318, Philadelphia, Pennsylvania, USA. Association for Computational Linguistics.

\bibitem[{Paszke et~al.(2019)Paszke, Gross, Massa, Lerer, Bradbury, Chanan, Killeen, Lin, Gimelshein, Antiga et~al.}]{paszke2019pytorch}
Adam Paszke, Sam Gross, Francisco Massa, Adam Lerer, James Bradbury, Gregory Chanan, Trevor Killeen, Zeming Lin, Natalia Gimelshein, Luca Antiga, and 1 others. 2019.
\newblock Pytorch: An imperative style, high-performance deep learning library.
\newblock \emph{Advances in neural information processing systems}, 32.

\bibitem[{Periti et~al.(2024)Periti, Alfter, and Tahmasebi}]{periti-etal-2024-automatically}
Francesco Periti, David Alfter, and Nina Tahmasebi. 2024.
\newblock \href {https://aclanthology.org/2024.emnlp-main.776} {Automatically generated definitions and their utility for modeling word meaning}.
\newblock In \emph{Proceedings of the 2024 Conference on Empirical Methods in Natural Language Processing}, pages 14008--14026, Miami, Florida, USA. Association for Computational Linguistics.

\bibitem[{Petridis et~al.(2024)Petridis, Wedin, Yuan, Wexler, and Thain}]{petridis2024constitutionalexperts}
Savvas Petridis, Ben Wedin, Ann Yuan, James Wexler, and Nithum Thain. 2024.
\newblock \href {https://doi.org/10.18653/v1/2024.acl-short.52} {{C}onstitutional{E}xperts: Training a mixture of principle-based prompts}.
\newblock In \emph{Proceedings of the 62nd Annual Meeting of the Association for Computational Linguistics (Volume 2: Short Papers)}, pages 574--582, Bangkok, Thailand. Association for Computational Linguistics.

\bibitem[{Pham et~al.(2023)Pham, Kim, Mukherjee, Woodruff, Poczos, and Hassan}]{pham-etal-2023-task}
Hai Pham, Young~Jin Kim, Subhabrata Mukherjee, David~P. Woodruff, Barnabas Poczos, and Hany Hassan. 2023.
\newblock \href {https://doi.org/10.18653/v1/2023.mrl-1.13} {Task-based {M}o{E} for multitask multilingual machine translation}.
\newblock In \emph{Proceedings of the 3rd Workshop on Multi-lingual Representation Learning (MRL)}, pages 164--172, Singapore. Association for Computational Linguistics.

\bibitem[{Pillutla et~al.(2021)Pillutla, Swayamdipta, Zellers, Thickstun, Welleck, Choi, and Harchaoui}]{pillutla2021mauve}
Krishna Pillutla, Swabha Swayamdipta, Rowan Zellers, John Thickstun, Sean Welleck, Yejin Choi, and Zaid Harchaoui. 2021.
\newblock Mauve: Measuring the gap between neural text and human text using divergence frontiers.
\newblock \emph{Advances in Neural Information Processing Systems}, 34:4816--4828.

\bibitem[{Pustejovsky and Boguraev(1993)}]{PUSTEJOVSKY1993193}
James Pustejovsky and Branimir Boguraev. 1993.
\newblock \href {https://doi.org/10.1016/0004-3702(93)90017-6} {Lexical knowledge representation and natural language processing}.
\newblock \emph{Artificial Intelligence}, 63(1):193--223.

\bibitem[{Rajbhandari et~al.(2020)Rajbhandari, Rasley, Ruwase, and He}]{rajbhandari2020zero}
Samyam Rajbhandari, Jeff Rasley, Olatunji Ruwase, and Yuxiong He. 2020.
\newblock Zero: Memory optimizations toward training trillion parameter models.
\newblock In \emph{SC20: International Conference for High Performance Computing, Networking, Storage and Analysis}, pages 1--16. IEEE.

\bibitem[{Reid et~al.(2024)Reid, Savinov, Teplyashin, Lepikhin, Lillicrap, Alayrac, Soricut, Lazaridou, Firat, Schrittwieser et~al.}]{reid2024gemini}
Machel Reid, Nikolay Savinov, Denis Teplyashin, Dmitry Lepikhin, Timothy Lillicrap, Jean-baptiste Alayrac, Radu Soricut, Angeliki Lazaridou, Orhan Firat, Julian Schrittwieser, and 1 others. 2024.
\newblock Gemini 1.5: Unlocking multimodal understanding across millions of tokens of context.
\newblock \emph{arXiv preprint arXiv:2403.05530}.

\bibitem[{Shao et~al.(2024)Shao, Dai, Guo, Liu, and Wang}]{Shao2024DeepSeekV2AS}
Zhihong Shao, Damai Dai, Daya Guo, Bo~Liu, and Zihan Wang. 2024.
\newblock \href {https://api.semanticscholar.org/CorpusID:269613809} {Deepseek-v2: A strong, economical, and efficient mixture-of-experts language model}.
\newblock \emph{ArXiv}, abs/2405.04434.

\bibitem[{Shazeer et~al.(2017)Shazeer, Mirhoseini, Maziarz, Davis, Le, Hinton, and Dean}]{shazeer2017}
Noam Shazeer, *Azalia Mirhoseini, *Krzysztof Maziarz, Andy Davis, Quoc Le, Geoffrey Hinton, and Jeff Dean. 2017.
\newblock \href {https://openreview.net/forum?id=B1ckMDqlg} {Outrageously large neural networks: The sparsely-gated mixture-of-experts layer}.
\newblock In \emph{International Conference on Learning Representations}.

\bibitem[{Shi et~al.(2024)Shi, Yang, Wu, Aitchison, Yilmaz, and Lipani}]{shi2024instruction}
Zhengyan Shi, Adam~X Yang, Bin Wu, Laurence Aitchison, Emine Yilmaz, and Aldo Lipani. 2024.
\newblock Instruction tuning with loss over instructions.
\newblock \emph{arXiv preprint arXiv:2405.14394}.

\bibitem[{Sparck~Jones(1972)}]{sparck1972statistical}
Karen Sparck~Jones. 1972.
\newblock A statistical interpretation of term specificity and its application in retrieval.
\newblock \emph{Journal of documentation}, 28(1):11--21.

\bibitem[{Srivastava et~al.(2014)Srivastava, Hinton, Krizhevsky, Sutskever, and Salakhutdinov}]{srivastava2014dropout}
Nitish Srivastava, Geoffrey Hinton, Alex Krizhevsky, Ilya Sutskever, and Ruslan Salakhutdinov. 2014.
\newblock Dropout: a simple way to prevent neural networks from overfitting.
\newblock \emph{The journal of machine learning research}, 15(1):1929--1958.

\bibitem[{Stiennon et~al.(2020)Stiennon, Ouyang, Wu, Ziegler, Lowe, Voss, Radford, Amodei, and Christiano}]{NEURIPS2020_1f89885d}
Nisan Stiennon, Long Ouyang, Jeffrey Wu, Daniel Ziegler, Ryan Lowe, Chelsea Voss, Alec Radford, Dario Amodei, and Paul~F Christiano. 2020.
\newblock \href {https://proceedings.neurips.cc/paper_files/paper/2020/file/1f89885d556929e98d3ef9b86448f951-Paper.pdf} {Learning to summarize with human feedback}.
\newblock In \emph{Advances in Neural Information Processing Systems}, volume~33, pages 3008--3021. Curran Associates, Inc.

\bibitem[{Sukhbaatar et~al.(2024)Sukhbaatar, Golovneva, Sharma, Xu, Lin, Roziere, Kahn, Li, tau Yih, Weston, and Li}]{sukhbaatar2024branch}
Sainbayar Sukhbaatar, Olga Golovneva, Vasu Sharma, Hu~Xu, Xi~Victoria Lin, Baptiste Roziere, Jacob Kahn, Shang-Wen Li, Wen tau Yih, Jason~E Weston, and Xian Li. 2024.
\newblock \href {https://openreview.net/forum?id=nqLAuMOF6n} {Branch-train-mix: Mixing expert {LLM}s into a mixture-of-experts {LLM}}.
\newblock In \emph{First Conference on Language Modeling}.

\bibitem[{Wolf et~al.(2020)Wolf, Debut, Sanh, Chaumond, Delangue, Moi, Cistac, Rault, Louf, Funtowicz et~al.}]{wolf2020transformers}
Thomas Wolf, Lysandre Debut, Victor Sanh, Julien Chaumond, Clement Delangue, Anthony Moi, Pierric Cistac, Tim Rault, R{\'e}mi Louf, Morgan Funtowicz, and 1 others. 2020.
\newblock Transformers: State-of-the-art natural language processing.
\newblock In \emph{Proceedings of the 2020 conference on empirical methods in natural language processing: system demonstrations}, pages 38--45.

\bibitem[{Wu et~al.(2023)Wu, Wang, Ye, Wu, Feng, Xu, and Qiao}]{wu-etal-2023-openicl}
Zhenyu Wu, Yaoxiang Wang, Jiacheng Ye, Zhiyong Wu, Jiangtao Feng, Jingjing Xu, and Yu~Qiao. 2023.
\newblock \href {https://doi.org/10.18653/v1/2023.acl-demo.47} {{O}pen{ICL}: An open-source framework for in-context learning}.
\newblock In \emph{Proceedings of the 61st Annual Meeting of the Association for Computational Linguistics (Volume 3: System Demonstrations)}, pages 489--498, Toronto, Canada. Association for Computational Linguistics.

\bibitem[{Yang et~al.(2024)Yang, Jimenez, Wettig, Lieret, Yao, Narasimhan, and Press}]{yang2024sweagent}
John Yang, Carlos~E Jimenez, Alexander Wettig, Kilian Lieret, Shunyu Yao, Karthik~R Narasimhan, and Ofir Press. 2024.
\newblock \href {https://openreview.net/forum?id=mXpq6ut8J3} {{SWE}-agent: Agent-computer interfaces enable automated software engineering}.
\newblock In \emph{The Thirty-eighth Annual Conference on Neural Information Processing Systems}.

\bibitem[{Yin and Skiena(2023)}]{yin2023word}
Yunting Yin and Steven Skiena. 2023.
\newblock Word definitions from large language models.
\newblock \emph{arXiv preprint arXiv:2311.06362}.

\bibitem[{Zhang et~al.(2022)Zhang, Li, Yang, and Li}]{zhang2022fine}
Hengyuan Zhang, Dawei Li, Shiping Yang, and Yanran Li. 2022.
\newblock Fine-grained contrastive learning for definition generation.
\newblock In \emph{Proceedings of the 2nd Conference of the Asia-Pacific Chapter of the Association for Computational Linguistics and the 12th International Joint Conference on Natural Language Processing (Volume 1: Long Papers)}, pages 1001--1012.

\bibitem[{Zhang et~al.(2023)Zhang, Chen, Wang, Jiang, Li, Wang, and Cao}]{zhang2023exploiting}
Linhan Zhang, Qian Chen, Wen Wang, Yuxin Jiang, Bing Li, Wei Wang, and Xin Cao. 2023.
\newblock Exploiting correlations between contexts and definitions with multiple definition modeling.
\newblock \emph{arXiv preprint arXiv:2305.14717}.

\bibitem[{Zhang et~al.(2019)Zhang, Kishore, Wu, Weinberger, and Artzi}]{zhang2019bertscore}
Tianyi Zhang, Varsha Kishore, Felix Wu, Kilian~Q Weinberger, and Yoav Artzi. 2019.
\newblock Bertscore: Evaluating text generation with bert.
\newblock In \emph{International Conference on Learning Representations}.

\bibitem[{Zhao et~al.(2019)Zhao, Peyrard, Liu, Gao, Meyer, and Eger}]{zhao2019moverscore}
Wei Zhao, Maxime Peyrard, Fei Liu, Yang Gao, Christian~M Meyer, and Steffen Eger. 2019.
\newblock Moverscore: Text generation evaluating with contextualized embeddings and earth mover distance.
\newblock In \emph{Proceedings of the 2019 Conference on Empirical Methods in Natural Language Processing and the 9th International Joint Conference on Natural Language Processing (EMNLP-IJCNLP)}, pages 563--578.

\bibitem[{Zhou et~al.(2025)Zhou, Karamanolakis, Soto, Rumshisky, Kulkarni, Huang, Ai, and Lu}]{zhou-etal-2025-mergeme}
Yuhang Zhou, Giannis Karamanolakis, Victor Soto, Anna Rumshisky, Mayank Kulkarni, Furong Huang, Wei Ai, and Jianhua Lu. 2025.
\newblock \href {https://aclanthology.org/2025.naacl-long.117/} {{M}erge{ME}: Model merging techniques for homogeneous and heterogeneous {M}o{E}s}.
\newblock In \emph{Proceedings of the 2025 Conference of the Nations of the Americas Chapter of the Association for Computational Linguistics: Human Language Technologies (Volume 1: Long Papers)}, pages 2315--2328, Albuquerque, New Mexico. Association for Computational Linguistics.

\bibitem[{Zhu et~al.(2024)Zhu, Qu, Dong, Ruan, Tong, He, and Cheng}]{zhu-etal-2024-llama}
Tong Zhu, Xiaoye Qu, Daize Dong, Jiacheng Ruan, Jingqi Tong, Conghui He, and Yu~Cheng. 2024.
\newblock \href {https://doi.org/10.18653/v1/2024.emnlp-main.890} {{LL}a{MA}-{M}o{E}: Building mixture-of-experts from {LL}a{MA} with continual pre-training}.
\newblock In \emph{Proceedings of the 2024 Conference on Empirical Methods in Natural Language Processing}, pages 15913--15923, Miami, Florida, USA. Association for Computational Linguistics.

\bibitem[{Zoph et~al.(2022)Zoph, Bello, Kumar, Du, Huang, Dean, Shazeer, and Fedus}]{zoph2022st}
Barret Zoph, Irwan Bello, Sameer Kumar, Nan Du, Yanping Huang, Jeff Dean, Noam Shazeer, and William Fedus. 2022.
\newblock St-moe: Designing stable and transferable sparse expert models.
\newblock \emph{arXiv preprint arXiv:2202.08906}.

\end{thebibliography}

\appendix
\section{Additional Experiment Details}
\label{sec:additional-experiment-results}

This is a section in the appendix. Introduce dataset components, hyperparameter settings, and other experimental details.
\paragraph{Data Processing.}

Raw 3D-EX (see fig. \ref{fig:3d-ex-components}) consists of ten lexicon sources of $\mathbb{<}t, c, d\mathbb{>}$ triplets, we use the word-level split on each of the sources to train, validate and test our models in this paper. We developed the following steps to undergo the preprocessing procedure for the raw 3D-EX dataset.

\begin{itemize}
    \item We discard instances that do not meet any of the following conditions: \ding{172} \textsc{Term} must be of string type, \ding{173} \textsc{Definition} must be of string type, \ding{174} \textsc{Example} must not be empty, and \ding{175} \textsc{Dataset\_Name} must not be empty.
    \item To enhance the model's ability to interpret words in various contexts, we split the sample entries with multiple example contexts into separate data instances for each context. This approach increases the number of samples the model sees during training.
\end{itemize}
\begin{center}
    \begin{tikzpicture}[scale=1.4]
      \foreach \frac/\name/\col [
        count=\i,
        remember=\angb as \anga (initially -10), 
        evaluate={\angm=\anga-\frac*1.8;  
                  \angb=\anga-\frac*3.6;  
                  \exp=\R*max(0.02,0.4*(1-\frac/100)^15); 
                  \r=\exp+\R*max(0.5,(0.83-\frac/100));} 
      ] in { 
           0.02/\text{Hei++}   /myyellow,        
           2.14/\text{CODWOE}       /mydarkestblue,    
           1.45/\text{WordNet}          /mycyan,   
          4.68/\text{Webster's Dict.}      /myorange, 
          4.76/Z\text{Urban}    /mybrown, 
          31.32/\text{Wikipedia}      /mygreen,  
          24.6/\text{CHA}         /myred,    
          5.44/\text{Sci-definition}    /mypink,   
          4.65/\text{Wiktionary}     /myblue, 
          20.95/\text{MultiRD}         /mypurple
      }{
        \coordinate (P\i) at (\angm:\exp+\R);
        \draw[slice=\col,line width={\frac>2?0.6:0.2},shadow opacity=100]
          (\angm:\exp) --++ (\anga:\R) arc(\anga:\angb:\R) -- cycle;
        \ifdim \frac pt > 9pt 
            \node[white,align=center]
            at (\angm:\r) {\boldlabel{$\name$}{\frac}};
        \else \ifdim \frac pt > 6pt 
          \node[white,scale={\frac>4?0.9:0.8}]
            at (\angm:\r) {\contour{\col!70!black!90}{\bm{$\frac\mathbf{\%}$}}};
          \node[\col!80!black,anchor=190+\angm,inner sep=4]
            at (P\i) {\bf\bm{$\name$}};
        \fi \fi
      }
      \draw[pin,myyellow!95!black]
        (P1)++(-30:0.04) --++ (-22:0.15)
        node[anchor=180,inner sep=2]
          {\bm{$\textbf{Hei++}$} \footnotesize$\mathbf{0.02\%}$};
    
      \draw[pin,mydarkestblue]
        (P2)++(-35:0.04) --++ (-33:0.15)
        node[anchor=170,inner sep=2]
          {\bm{$\textbf{CODWOE}$} \footnotesize$\mathbf{2.14\%}$};
    
      \draw[pin,mycyan]
        (P3)++(-60:0.04) --++ (-58:0.15)
        node[anchor=170,inner sep=2]
          {\bm{$\textbf{WordNet}$} \footnotesize$\mathbf{1.45\%}$};
    
      \draw[pin,myorange]
      (P4)++(-55:0.04) --++ (-80:0.15)
      node[anchor=150,inner sep=2,text width=2cm,align=center]
        {\bm{$\textbf{Webster's Dict.}$}\vspace{-0.4em}\ \footnotesize$\mathbf{4.68\%}$};
    
      \draw[pin,mybrown]
        (P5)++(-30:0.04) --++ (-105:0.15)
        node[anchor=40,inner sep=1.8]
          {\bm{$\textbf{Urban}$} \footnotesize$\mathbf{4.76\%}$};
    
      \draw[pin,mypink!85!black]
        (P8)++(110:0.03) --++ (165:0.15)
        node[anchor=-15,inner sep=2]
          {\bm{$\textbf{Sci-definition}$} \footnotesize$\mathbf{5.44\%}$};
    
      \draw[pin,myblue]
        (P9)++(-260:0.04) --++ (10:0.15)
        node[anchor=180,inner sep=2]
          {\bm{$\textbf{Wiktionary}$} \footnotesize$\mathbf{4.65\%}$};
    
      \node[above,scale=1.1] at (0,1.5) {\bf ~~~~~~~~3D-EX Constituents Dist. (\%)};
    \end{tikzpicture}
    
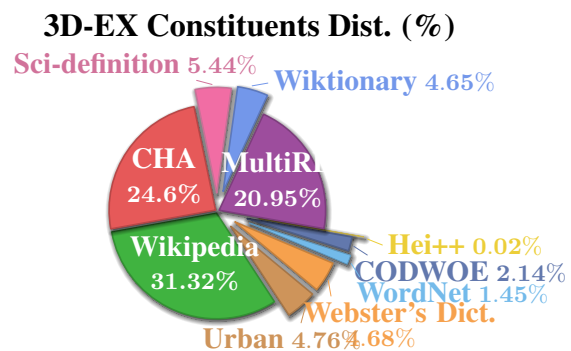
\captionof{figure}{3D-EX constituents distribution.}
    \label{fig:3d-ex-components}
\end{center}
In addition, we observed many examples in the existing datasets that share the same term-context pair but with different definitions, which may cause negative effects on model learning if there exist many semantics-divergent examples. To summarize and display the potential impacts, we report the salient statistics about this finding of these datasets shown in the following Table \ref{tab:divergent-examples-statistics}.
\begin{table}[ht]
\begin{center}
\addtolength{\tabcolsep}{-3.5pt}
\begin{tabular}{ccccc}
\toprule
\textbf{Dataset} & \textbf{Split} & \textbf{\# All} & \textbf{\# Div.} & \textbf{\% Div. / All}
\\
\midrule
\multirow{3}{*}{WordNet}
& $\mathcal{S}_\text{train}$ & 13,883 & 2,723 & \hlcellrrrr 19.61
\\
& $\mathcal{S}_\text{valid}$ & 1,752 & 368 & \hlcellrrrr 21.00
\\
& $\mathcal{S}_\text{test}$ & 1,775 & 333 & \hlcellrrrr 18.76
\\
\midrule
\multirow{3}{*}{Oxford}
& $\mathcal{S}_\text{train}$ & 82,479 & 34 & \hlcellr 0.04
\\
& $\mathcal{S}_\text{valid}$ & 10,285 & 2 & \hlcellr 0.02
\\
& $\mathcal{S}_\text{test}$ & 10,306 & 0 & 0.00
\\
\midrule
\multirow{3}{*}{Wikipedia}
& $\mathcal{S}_\text{train}$ & 887,455 & 186 & \hlcellr 0.02
\\
& $\mathcal{S}_\text{valid}$ & 44,003 & 16 & \hlcellr 0.04
\\
& $\mathcal{S}_\text{test}$ & 57,232 & 14 & \hlcellr 0.02
\\
\midrule
\multirow{3}{*}{Urban}
& $\mathcal{S}_\text{train}$ & 411,382 & 1,424 & \hlcellrr 0.35
\\
& $\mathcal{S}_\text{valid}$ & 57,883 & 152 & \hlcellrr 0.26
\\
& $\mathcal{S}_\text{test}$ & 38,371 & 122 & \hlcellrr 0.32
\\
\midrule
\multirow{3}{*}{3D-EX}
& $\mathcal{S}_\text{train}$ & 1,309,312 & 35,632 & \hlcellrrr 2.72
\\
& $\mathcal{S}_\text{valid}$ & 513,789 & 12,551 & \hlcellrrr 2.44
\\
& $\mathcal{S}_\text{test}$ & 450,078 & 7,599 & \hlcellrrr 1.69
\\
\bottomrule
\end{tabular}
\end{center}
\caption{Divergent examples statistics of each dataset. \textbf{\# All}: number of all examples; \textbf{\# Div.}: number of all divergent examples; \textbf{\% Div. / All}: ratio of divergent examples in all examples.}
\label{tab:divergent-examples-statistics}
\vspace{-1.2em}
\end{table}

\paragraph{Clustering Setup.}
Compared with \citet{gururangan2023scaling}, we consider to mine the intrinsit semantic meaning of term associated with their context, instead of using lexical statistics clustering method, like TF-IDF. We argue that the method building on dense semantic clustering would help upcycling models to learn specialized sense interpretation-oriented experts, towards robust system for definition modeling. We run k-means++ clustering of the Elkan variation method with $1,000$ max iteration, $1e^{-8}$ tolerance of convergence, and a fixed seed of $42$. Considering the computation and memory bounds, we first use $4$ as the number of clusters to form and the number of centroids to generate. We further ablate this factor in the section \S\ref{subsec:ablation_study}.

\paragraph{Training Details.}
\ourmethod was trained for 3 epochs with a global batch size of 8,192 tokens (gradient accumulation 1, batch size per device 8, max sequence length 128) on 8 $\times$ H100-PCIe-80GB GPUs and a learning rate of 1e-6, minimum learning rate of 3e-7 with a cosine annealing scheduler, as well as the warm-up steps with 6\% ratio of the total training steps. We used a global dropout of 0.2 \cite{srivastava2014dropout} and a weight decay of 0.1 with AdamW optimizor \cite{loshchilov2018decoupled}, and performed early stopping to obtain the best model by the highest validation bleu.

Moreover, We run three times for each training setup to report the mean results and their standard deviation of metrics, with seed $s_i \in \left\{21, 42, 84\right\}$, respectively. We use Hugging Face Transformers \cite{wolf2020transformers} and Pytorch \cite{paszke2019pytorch} to develop the training pipeline.

We run the branch training on each cluster of data points obtained from the clustering results. As depicted in tab. \ref{tab:training-hyperparams}, We set up the following hyper-parameters to train \ourmethod and vanilla fine-tuned \textsc{Llama-3-8B} models in this paper. We used the standard negative log-likelihood (NLL) loss to train \ourmethod. Contrary to \citet{shi2024instruction}, to avoid the loss of the input sequence tokens overshadowing the actual output token loss, the loss is only computed over the result tokens (Eq. \ref{eq:loss}), limiting the potential to overfit to the input prompt and context. This loss calculation method resulted in faster training and robuster results overall.

Given a definition generation problem $p(c, t)$ and its golden reference $d$, we define a outcome reward model as the following: ORM ($P \times D \rightarrow \mathbb{R}$) assigns a single value to $s$ to indicate whether predicted $\hat{d}$ is correct. Given a specific dataset $\mathcal{D}$, we follow \citet{cobbe2021training} to use a negative log-likelihood loss (Eq. \ref{eq:orm_objective}) to frame the reward modeling as a binary classification objective.
\begin{equation}\label{eq:orm_objective}
    \mathcal{L}_{\mathrm{ORM}} =-\log \sigma\left(r_\phi(x, y_w)-r_\phi(x, y_l)\right)
\end{equation}
Where $y_w$ is the preferred generation (i.e., chosen response) and $y_l$ is the alternate generation (i.e., rejected response) conditioned on the input $x:=p(c, t)$. To train a ORM built on training set, we leverage the golden reference $d$ as the preferred definition $y_w$ and one of the model generations as the alternate definition $y_l$ to express preferences for each $x$, denoted as $y_w\succ y_l \mid x$, where $y_w$ and $y_l$ denotes the preferred and dispreferred completion, respectively. $\sigma$ is the sigmoid function and $r_\phi(\cdot, \cdot)$ represents the parameterized reward function for the concatenated input $x$ and generation $y_*$. To enhance computing efficiency, we employ the ratio of $1:32$ to conduct repeated sampling and rerank the generations by their log-likelihood (aka. confidence) to acquire the top-eight items as a candidate set of alternate generations for each input $x$.

\paragraph{Inference Setup.}
As shown in Table \ref{table:main_results}, for each setting in ``Zero-shot'', ``BoN-Oracle'', and ``BoN-ORM'', we orchestrate three separate runs for each setting, using the same decoding parameters but with different random seeds to ensure robustness and consistency in the results. Specifically, for the models \textsc{LM-Lexicon-Dense} and \textsc{LM-Lexicon-MoE}, specifically, we use the temperature of $0.6$, $top\text{-}k$ of $50$, $top\text{-}p$ of $0.9$, and repetition penalty of $1.05$, ensuring uniformity across all evaluations.

For all benchmarks included in our test, as the number of samples increases, the coverage metric corresponds to the use of an oracle verifier. This verifier checks which fraction of DM problems in the test set can be approximated using any of the samples that were generated to be as similar as possible to the ground truth. The selection of the most similar generation is achieved through an iterative comparison with the golden definition, ensuring a robust matching process. In the case of the oracle verification process by the oracle verifier, we validate whether any output chosen prediction is the most similar by comparing it with golden references of the sample in the test set. In contrast, for the verification process of ORM verifier, the selection of the most similar generation is then performed solely by the ORM verifier itself, without relying on external feedback, ground-truth comparison, or oracle input.

\paragraph{Miscellaneous.}
We developed our MoE language modeling codebase based on \citet{mergoo2024} and implemented several routing policies and proposed MoE architectures. Aiming at more efficent evlauation, we follow \cite{huang-etal-2021-definition} and refactor their implementation with concurrent metrics computation to boost the inference procedure in large models, please see the details in our released code.

\section{Carbon Footprint}\label{sec:training-inference-cost}
The cost of fine-tuning LLM is lower than that of pre-training them. Nevertheless, we think it is critical to quantify and record the environmental consequences of our research. Table \ref{tab:carbon-footprint} lists the materials required for a single run, which is conducted using our own infrastructure. We calculate the carbon footprint estimation using a carbon intensity of 0.141 kg/kWh and 700W consumption per GPU\footnote{Statistics: \url{https://app.electricitymaps.com/map}.}.

\begin{table}[ht]
\centering
\resizebox{1\linewidth}{!}{
    \begin{tabular}{lcccc}
    \toprule
    \textbf{Model} & \textbf{Hardware} & \textbf{FLOPs} & \textbf{Time (h)} & \textbf{CO2eq (kg)} \\
    \midrule
    \textsc{\LogoMiddle\ LM-Lexicon-Dense} & 8$\times$H100 & $4.2e^{18}$ & $36.4$ & $11.4$ \\
    \textsc{\LogoMiddle\ LM-Lexicon-MoE} & 8$\times$H100 & $5.4e^{18}$ & $32.8$ & $14.6$ \\
    \bottomrule
    \end{tabular}
}
\caption{Details about the training required resources.}
\label{tab:carbon-footprint}
\end{table}

\section{Additional Evaluation Results}
\label{sec:additional-evaluation-results}

\subsection{Data Clustering Results}
\label{subsec:data_clustering_results}

\begin{table}[ht]
    \centering
    \begin{tabular}{lc}
        \toprule 
        \textbf{Cluster $C_i$} & \textbf{Distance}$_\text{intra-cluster}$ $\downarrow$ \\
        \midrule 
        $C_0$ (Adjective) & 0.176 \\
        $C_1$ (Scientific) & 0.168 \\
        $C_2$ (Proper Noun) & 0.173 \\
        $C_3$ (Person Name) & 0.185 \\
        \midrule 
        Average & 0.175 \\
        \bottomrule 
    \end{tabular}
    \caption{Intra-cluster Distances (\textit{i.e.,} the cluster cohesion)}.
    \label{tab:intra_cluster_distances}
\end{table}

\begin{table}[ht]
    \centering
    \begin{tabular}{lc}
        \toprule 
        \textbf{Cluster ($C_i$, $C_j$)} & \textbf{Distance}$_\text{inter-cluster}$ $\uparrow$ \\
        \midrule 
        $C_0$, $C_1$ & 0.694 \\
        $C_0$, $C_2$ & 0.713 \\
        $C_0$, $C_3$ & 0.765 \\
        $C_1$, $C_2$ & 0.681 \\
        $C_1$, $C_3$ & 0.707 \\
        $C_2$, $C_3$ & 0.720 \\
        \midrule 
        Average & 0.713 \\
        \bottomrule 
    \end{tabular}
    \caption{Inter-cluster Distances (\textit{i.e.,} the cluster separation): $C_0$ denotes the domain of ``Adjective'', $C_1$ denotes the domain of ``Scientific'', $C_2$ denotes the domain of ``Proper Noun'', and $C_3$ denotes the domain of ``Person Name''.}
     \label{tab:inter_cluster_distances}
\end{table}

We show the clustering results including cluster cohesion and cluster separation in the following Table \ref{tab:intra_cluster_distances} and \ref{tab:inter_cluster_distances}, respectively.

\subsection{In-Context Learning Evaluation}
\begin{figure*}[ht]
  \centering
  \resizebox{0.95\linewidth}{!}{\includegraphics{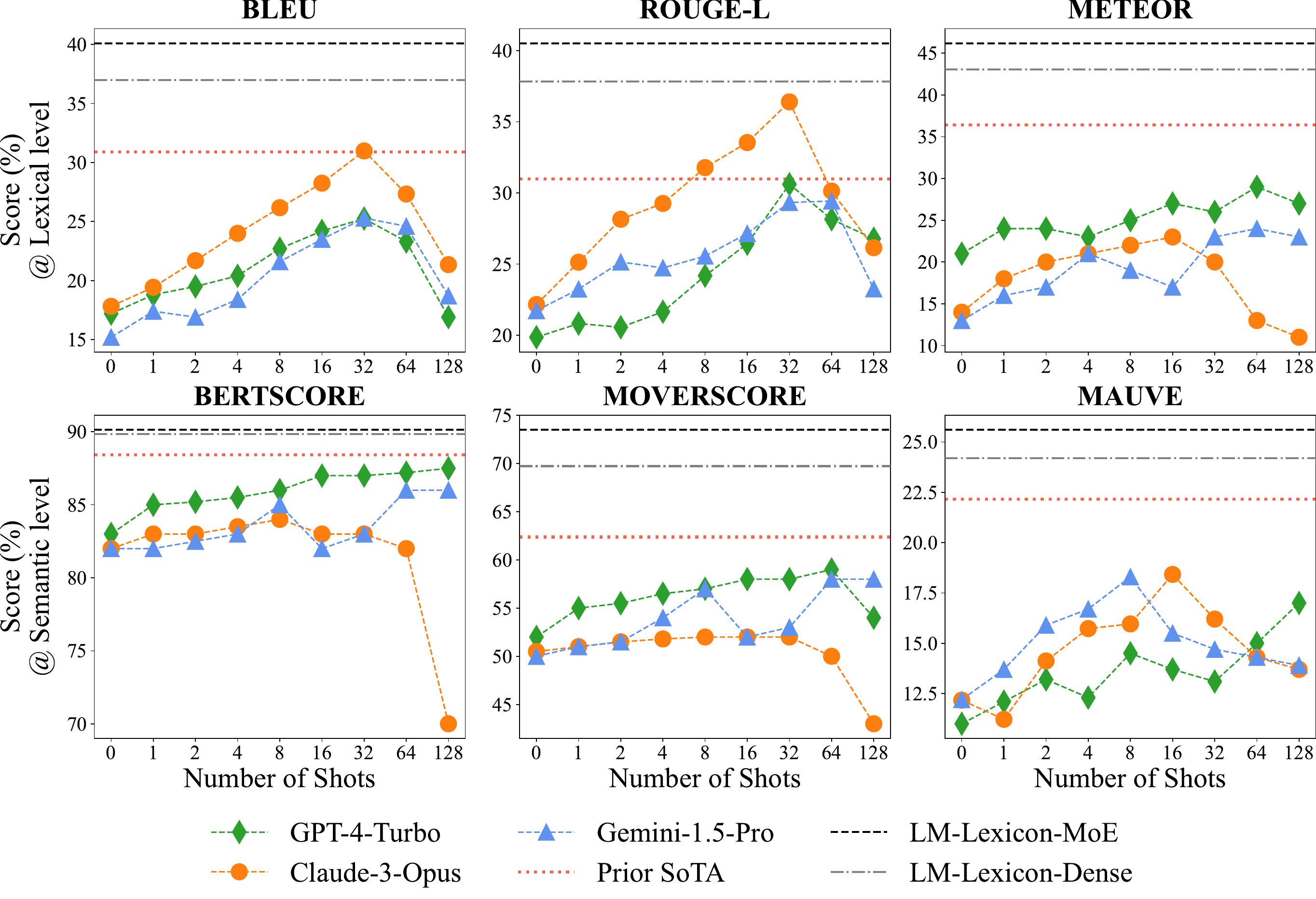}}
  \caption{Scaling the in-context learning results of frontier causal LMs on WordNet with $k$-shot demonstrations, where $k$ scales logarithmically from $0$ to $128$. Prior SoTA denotes the Rerank-T5 proposed by \citet{huang-etal-2021-definition}.}
  \label{fig:icl-scaling-results}
\end{figure*}

We show the scaling in-context learning experimental results as shown in Figure. \ref{fig:icl-scaling-results}.

\subsection{Generation Examples of \ourmethod}
As depicted in Figure \ref{fig:data-example-cluster-1}, \ref{fig:data-example-cluster-2}, \ref{fig:data-example-cluster-3}, and \ref{fig:data-example-cluster-4}, we provide a cherry-picked example for each domain cluster as shown in Figure \ref{fig:3d-ex-visualized-cluster} in definition modeling.

\begin{figure}[H]
    \begin{framed}
        \textbf{Cluster-1 Example:}\\
        \small
        \textbf{[Term]} \textcolor[HTML]{3078BE}{\textbf{Combtooth Blenny}} \\
        \textbf{[Query]} \textcolor[HTML]{3078BE}{``the crested blenny is a species of \textbf{Combtooth Blenny} found around New South Wales, Australia, ...'' \textcolor{black}{What is the definition of} ``\textbf{Combtooth Blenny}''?} \\
        \textbf{[Source]} Wikipedia \\
        \textbf{[Reference]} \textcolor[HTML]{3078BE}{\textbf{Combtooth Blenny}}: perciform marine fish of the family blenniidae.
        \normalsize
    \end{framed}
    \caption{Example of $\mathcal{C}_1$ (proper noun) from 3D-EX.}
    \label{fig:data-example-cluster-1}
\end{figure}

\begin{figure}[H]
    \begin{framed}
        \textbf{Cluster-2 Example:}\\
        \small
        \textbf{[Term]} \textcolor[HTML]{3078BE}{\textbf{brave}} \\
        \textbf{[Query]} \textcolor[HTML]{3078BE}{``familiarity with danger makes a \textbf{brave} man braver but less daring - herman melville ...'' \textcolor{black}{What is the definition of} ``\textbf{brave}''?} \\
        \textbf{[Source]} WordNet \\
        \textbf{[Reference]} \textcolor[HTML]{3078BE}{\textbf{brave}}: possessing or displaying courage; able to deal with danger or fear without flinching.
        \normalsize
    \end{framed}
    \caption{Example of $\mathcal{C}_2$ (adjective) from 3D-EX.}
    \label{fig:data-example-cluster-2}
\end{figure}

\begin{figure}[H]
    \begin{framed}
        \textbf{Cluster-3 Example:}\\
        \small
        \textbf{[Term]} \textcolor[HTML]{3078BE}{\textbf{Michael Maclennan}} \\
        \textbf{[Query]} \textcolor[HTML]{3078BE}{``Godiva's is a Canadian television comedy-drama series created by \textbf{Michael Maclennan} with Julia Keatley of Keatley Entertainment ...'' \textcolor{black}{What is the definition of} ``\textbf{Michael Maclennan}''?} \\
        \textbf{[Source]} Wikipedia \\
        \textbf{[Reference]} \textcolor[HTML]{3078BE}{\textbf{Michael Maclennan}}: Canadian playwright, screenwriter, and producer of television shows.
        \normalsize
    \end{framed}
    \caption{Example of $\mathcal{C}_3$ (person name) from 3D-EX.}
    \label{fig:data-example-cluster-3}
\end{figure}

\begin{figure}[H]
    \begin{framed}
        \textbf{Cluster-4 Example:}\\
        \small
        \textbf{[Term]} \textcolor[HTML]{3078BE}{\textbf{Lymphedema-distichiasis Syndrome}} \\
        \textbf{[Query]} \textcolor[HTML]{3078BE}{``two patients with \textbf{Lymphedema-distichiasis Syndrome} illustrate that both Milroy's ...'' \textcolor{black}{What is the definition of} ``\textbf{Lymphedema-distichiasis Syndrome}''?} \\ 
        \textbf{[Source]} Sci-definition \\
        \textbf{[Reference]} \textcolor[HTML]{3078BE}{\textbf{Lymphedema-distichiasis Syndrome}}: lymphedema distichiasis syndrome is a condition that affects the normal function of the lymphatic system.
        \normalsize
    \end{framed}
    \caption{Example of $\mathcal{C}_4$ (scentific) from 3D-EX.}
    \label{fig:data-example-cluster-4}
\end{figure}


\section{Human Evaluation Agreement}
\label{sec:annotators' agreement}
To assess the agreement among the annotators, we employed Fleiss's Kappa \cite{Fleiss_1971}, which is a statistical measurement to assess the reliability of the agreement between multiple raters. Fleiss's Kappa account for the possibility of agreement occurring by chance. It is calculated using the following formula:
\[
\kappa = \frac{P_o - P_e}{1 - P_e}
\]
where:
\begin{itemize}
  \item \( P_o \) is the observed agreement among the raters, and
  \item \( P_e \) is the expected agreement by chance.
\end{itemize}
Table \ref{tab:kappa_values_tab} presents Fleiss's Kappa coefficients for human evaluation agreement on each criterion and model.

\begin{table*}[ht]
\label{tab:kappa_values}
\centering
\small
\begin{tabular}{lccccc}
\toprule
\textbf{Criteria} & LM-Lexicon-MoE & LM-Lexicon-Dense & Claude-3-Opus & Gemini-1.5-Pro & GPT-4-Turbo \\
\midrule
Accuracy & 0.85 & 0.78 & 0.80 & 0.79 & 0.77 \\
Clarity & 0.82 & 0.75 & 0.78 & 0.76 & 0.74 \\
Conciseness & 0.80 & 0.74 & 0.77 & 0.75 & 0.73 \\
Context Appropriateness & 0.82 & 0.76 & 0.79 & 0.78 & 0.76 \\
Grammar \& Fluency & 0.78 & 0.76 & 0.75 & 0.74 & 0.73 \\
\bottomrule
\end{tabular}
\caption{Fleiss's Kappa coefficient statistics for the agreement of human evaluation on each criterion and model.}
\label{tab:kappa_values_tab}
\end{table*}
\newpage

\section{Comparison of Different Definitions}
\label{sec:representative_case}
The following is a representative case of the generated definitions from five models including three proprietary models, LM-Lexicon-Dense and LM-Lexicon-MoE:

\textbf{Word}: "coon”

\textbf{Context}: "I'll be a gone coon when the battle starts”

\textbf{Reference}: "an eccentric or undignified rustic” (from WordNet)

In the demonstration in Table \ref{tab:case}, definition generated by LM-Lexicon-MoE (our method) is most closely aligned with the reference definition's core meaning, which captures the derogatory connotation and reflects the "undignified" aspect. In contrast, definitions produced by the other models fail to capture fundamental aspects of the reference definition, for instance, Claude-3-Opus and GPT-4-Turbo completely deviate from the basic meaning of the target word in the context.

In general, there are some notable patterns:
\begin{itemize}
    \item Closed-source models (e.g., GPT-4 and Claude) generate overly verbose definitions, including redundant contextual information and multiple interpretations.
    
    \item These models tend to overemphasize derivative or secondary meanings, resulting in broad, general definitions that deviate from the core semantic meaning.
    
    \item Models trained on general corpora often struggle with uncommon word usages due to limited training examples, leading to inaccurate definitions or contextual misinterpretation.
\end{itemize}
This comparison demonstrates the superiority of LM-Lexicon-MoE's targeted, controlled approach over general-purpose language models for definition modeling.

\begin{table*}[t]
\centering
\setlength{\tabcolsep}{0pt}  
\begin{tabularx}{\textwidth}{X p{10cm}}  
\toprule
\textbf{Model} & \textbf{Generated Definition} \\
\midrule
Claude-3-Opus & The phrase gone coon means someone who is certainly defeated or about to die. \\
\midrule
Gemini-1.5-Pro & The word ``coon'' is a deeply offensive racial slur targeting people of African descent. It has a long and painful history rooted in racist caricatures and stereotypes. While it once appeared casually in entertainment and everyday language, it is now considered completely unacceptable to use. \\
\midrule
GPT-4-Turbo & ``Coon'' is a colloquial abbreviation for ``raccoon,'' a nocturnal mammal native to North America. ``Coon'' was sometimes used to refer to a person who is in deep trouble or doomed, often in the phrase ``a gone coon,'' meaning someone who is as good as dead or has no chance of survival. \\
\midrule
\LogoSmall\ LM-Lexicon-Dense (Ours) & A person who is afraid \\
\midrule
\LogoSmall\ LM-Lexicon-MoE (Ours) & A person who is deemed to be despicable or contemptible \\
\bottomrule
\end{tabularx}
\caption{Comparison of generated definition by models.}
\label{tab:case}
\label{tab:definitions}
\end{table*}

\section{Code for \LogoMiddle\ \ourmethod}
\label{sec:code}
We use the Alg.~\ref{code:merger} and~\ref{code:routers} provided below to train \ourmethod-\textsc{MoE} on the 3D-EX and the other four legacy datasets used in this paper. We exposed this Pytorch-style code as an implementation to extend our method to any potential domain. Additionally, to illustrate the differences and novelty of our method, we provide a comparison with the relative methods as shown in Table \ref{tab:method-comparison}.

\definecolor{codeblue}{rgb}{0.25,0.5,0.5}
\definecolor{codeblue2}{rgb}{0,0,1}
\lstset{
  backgroundcolor=\color{white},
  basicstyle=\fontsize{10pt}{10pt}\ttfamily\selectfont,
  columns=fullflexible,
  breaklines=true,
  captionpos=b,
  commentstyle=\fontsize{8pt}{8pt}\color{codeblue},
  keywordstyle=\fontsize{8pt}{8pt}\color{codeblue2},
}

\begin{algorithm*}[!ht]
\caption{\large Pytorch code for semantic experts merger.}
\begin{lstlisting}[language=Python]

def merge_semantic_experts(experts, router_layers):
    """
    Merge expert models into a unified model.

    Args:
        - experts (ModuleList): Experts to merge.
        - router_layers (ModuleList): Router layers.

    Returns:
        - state_dict (Dict[str, Tensor]): Merged model weights.
    """
    state_dict = dict()
    expert_nums = len(experts)
    count_total_router_layers = 0
    
    for idx, expert in enumerate(experts):
        # load each expert model
        model_id = expert["model_id"]
        model = load_base_model(model_id)
        
        if hasattr(model, "_tied_weights_keys"):
            tied_weights_keys.extend(model._tied_weights_keys)
            count_router_layers = 0
            count_averaged_layers = 0
            
        # iterate over all the layers of the model
        for layer_name, param in model.state_dict().items():
            is_merge_layer = True
            for router_layer in router_layers:
                if is_layer_suitable_for_router(router_layer, layer_name):
                    is_merge_layer = False
                    wb = layer_name.split(".")[-1]
                    new_layer_name = layer_name.split(f"{wb}")[0]
                    new_layer_name = f"{new_layer_name}experts.{ix}.{wb}"
                    assert new_layer_name not in state_dict
                    state_dict[new_layer_name] = param
                    count_total_router_layers += 1
                    count_router_layers += 1

            if is_merge_layer:
                # average the rest of layers by mean of weights
                prev_weight = state_dict.get(layer_name)
                
                if prev_weight is None:
                    prev_weight = torch.tensor(0)
                else:
                    if not prev_weight.shape == param.shape:
                        # adjust the shape of weight
                        prev_weight, param = shape_adjuster(
                            prev_weight, param, idx
                        )

                try:
                    # sometimes data is empty / non weights
                    state_dict[layer_name] = prev_weight + (param / expert_nums)
                except Exception as _:
                    print(layer_name, param)
                    state_dict[layer_name] = param

                count_averaged_layers += 1

    return state_dict
        
\end{lstlisting}
\label{code:merger}
\end{algorithm*}

\begin{algorithm*}[!ht]
\caption{\large Pytorch code for modeling \textsc{LM-Lexicon-MoE} Layer}
\begin{lstlisting}[language=Python]

class SemanticMoeLayer(nn.Module):
    def __init__(
        self,
        in_features: int,
        out_features: int,
        bias: bool,
        num_experts: int,
        num_experts_per_tok: int = 2,
        routing_policy: str,
    ):
        """Semantic Mixture-of-Experts Layer.

        Args:
            - in_features (int): Input Features
            - out_features (int): Output Features
            - bias (bool): Use bias or not.
            - num_experts (int): Total numbers of experts that Router Layer would handle
            - num_experts_per_tok (int): Number of active experts per token.
            - routing_policy (str): Routing Policy.
        """
        super().__init__()
        self.routing_policy = routing_policy
        if routing_policy == "token-level":
            # top-k token-level routing
            self.gate = nn.Linear(in_features, num_experts, bias=False)
            self.experts = nn.ModuleList(
                [nn.Linear(in_features, out_features, bias) for _ in range(num_experts)]
            )
            self.num_experts_per_tok = num_experts_per_tok
            self.in_features = in_features
            self.out_features = out_features
        elif routing_policy in ["soft-sequence-level", "hard-sequence-level"]:
            # soft/hard sequence-level routing
            self.gate = nn.Linear(in_features, num_experts, bias=False)
            self.num_experts = num_experts
            self.experts = nn.ModuleList(
                [nn.Linear(in_features, out_features) for _ in range(num_experts)]
            )
        elif routing_policy == "domain-level":
            # domain-level routing
            self.gate = nn.Linear(in_features, num_experts, bias=False)
            self.num_experts = num_experts
            self.experts = nn.ModuleList(
                [nn.Linear(in_features, out_features) for _ in range(num_experts)]
            )

    def forward(self, inputs: torch.Tensor, domain_labels: torch.Tensor):
        if self.routing_policy == "token-level":
            gate_logits = self.gate(inputs)
            weights, selected_experts = torch.topk(
                gate_logits, self.num_experts_per_tok
            )
            weights = F.softmax(weights, dim=2, dtype=torch.float).to(inputs.dtype)
            results = torch.zeros(
                (inputs.shape[0], inputs.shape[1], self.out_features),
                device=inputs.device,
                dtype=inputs.dtype,
            )
            
    # continue this table as below ...
            
\end{lstlisting}
\label{code:routers}
\end{algorithm*}
\newpage

\begin{algorithm*}[!ht]
\captionsetup{labelformat=empty}
\setlength{\floatsep}{0pt}
\begin{lstlisting}[language=Python]

    # continue the above table ...
    
            weights = weights.to(inputs.device)
            for ix, expert in enumerate(self.experts):
                batch_idx, tok_idx, expert_idx = torch.where(selected_experts == ix)
                results[batch_idx, tok_idx] += expert(
                    inputs[batch_idx, tok_idx]
                ) * weights[batch_idx, tok_idx, expert_idx].unsqueeze(-1)
        elif self.routing_policy == "soft-sequence-level":
            # soft sequence-level routing
            gate_logits = self.gate(inputs)
            gate_logits_mean = gate_logits.mean(dim=1)
            weights = F.softmax(gate_logits_mean, dim=-1)
            results = torch.zeros(
                (inputs.shape[0], inputs.shape[1], self.out_features),
                device=inputs.device,
                dtype=inputs.dtype,
            )
            for ix, expert in enumerate(self.experts):
                results += expert(inputs) * weights[:, ix].unsqueeze(-1)
        elif self.routing_policy == "hard-sequence-level":
            # hard sequence-level routing (only one selected expert is responsible for the entire sequence)
            gate_logits = self.gate(inputs)
            gate_logits_mean = gate_logits.mean(dim=1)
            _, selected_experts = torch.topk(gate_logits_mean, 1)
            results = torch.zeros(
                (inputs.shape[0], inputs.shape[1], self.out_features),
                device=inputs.device,
                dtype=inputs.dtype,
            )
            for ix, expert in enumerate(self.experts):
                results += expert(inputs) * (selected_experts == ix).float().unsqueeze(
                    -1
                )
        elif self.routing_policy == "domain-level":
            # domain-level routing (only one selected expert is responsible for the entire sequence)
            gate_logits = self.gate(inputs)
            results = torch.zeros(
                (inputs.shape[0], inputs.shape[1], self.out_features),
                device=inputs.device,
                dtype=inputs.dtype,
            )
            for ix, expert in enumerate(self.experts):
                results += expert(inputs) * (domain_labels == ix).float().unsqueeze(-1)

        return results
            
\end{lstlisting}
\label{code:routers-continue}
\end{algorithm*}
\newpage
\begin{table*}[!ht]
    \centering
    \small
    \begin{tabular}{lcccc}
    \toprule
    & \begin{tabular}{c} 
    \hyperlinkcite{shazeer2017}{\textsc{MoE (2017)}} \\
    (Vanilla)
    \end{tabular} & \begin{tabular}{c} 
    \hyperlinkcite{li2022branch}{BTM (2022)} \\
    (Merge)
    \end{tabular} & \begin{tabular}{c} 
    \hyperlinkcite{sukhbaatar2024branch}{BTX (2024)} \\
    (Linear router)
    \end{tabular} & \begin{tabular}{c} 
    \LogoSmall\ \ourmethod \\
    (Ours)
    \end{tabular} \\
    \midrule
    \makecell[l]{$\diamondsuit$ Dense experts are \\ trained independently (upcycling)} & \redcross & \greencheck & \greencheck & \greencheck \\
    \makecell[l]{$\diamondsuit$ Experts are specialized \\ in different domains} & \redcross & \greencheck & \greencheck & \greencheck \\
    \makecell[l]{$\diamondsuit$ Experts are chosen by \\ a learned router per input token} & \greencheck & \redcross & \greencheck & \greencheck \\
    \makecell[l]{$\diamondsuit$ Adaptive router via \\ domain-wise routing} & \redcross & \redcross & \redcross & \greencheck \\
    \makecell[l]{$\diamondsuit$ Semantic experts \\ adapted to diverse domains} & \redcross & \redcross & \redcross & \greencheck \\
    \bottomrule
\end{tabular}
\caption{A comprehensive comparison of the most relative sparse mixture-of-experts frameworks in recent years, including MoE (Vanilla), BTM (Merge), BTX (Linear Router), and \ourmethod. Our method demonstrates advancements in semantic-centric specialized expert and adaptability across domains.}
\label{tab:method-comparison}
\end{table*}
\begin{table*}[!ht]
    \centering
    \small
    \begin{tabular}{c}
      \toprule
      \textbf{Computing Infrastructure}\\8 \texttimes\ H100-80GB GPU (PCIe) \\ 
      \midrule \\
    \vspace{3mm}
    \begin{tabular}{ll}
\midrule \textbf{Hyperparameter} & \textbf{Assignment} \\
\midrule Base model & LM-Lexicon-Dense\\& (Llama-3-8B) \\
    Training strategy & \textsc{DS ZeRO-3} \\
    Epochs & 3 \\
    Global batch size & 524,288 tokens \\
    Max sequence length & 128 \\
    Max learning rate & $5\mathrm{e}-6$ \\
    Optimizer & AdamW \\
    Adam beta weights & $0.9,0.95$ \\
    Learning rate schedule & Cosine decay to 0 \\
    Weight decay & 0.01 \\
    Warm-up ratio & 10\% \\
    Gradient clipping & 1.0 \\
    Global dropout & 0.1 \\
    Random seeds & $\{21, 42, 84\}$ \\
\midrule
\end{tabular}
    \vspace{-3mm}
    \begin{tabular}{ll}
    \midrule \textbf{Hyperparameter} & \textbf{Assignment} \\
    \midrule Base model & LM-Lexicon-MoE\\& (4 \texttimes\ Llama-3-8B) \\
    Training strategy & \textsc{Naive PP} \\
    Epochs & 1 \\
    Global batch size & 131,072 tokens \\
    Max sequence length & 128 \\
    Max learning rate & $1\mathrm{e}-6$ \\
    Optimizer & AdamW \\
    Adam beta weights & $0.9,0.95$ \\
    Learning rate schedule & Cosine decay to 0 \\
    Weight decay & 0.01 \\
    Warm-up ratio & 10\% \\
    Gradient clipping & 1.0 \\
    Global dropout & 0.1 \\
    Random seeds & $\{21, 42, 84\}$ \\
    \midrule
    \end{tabular}
    \vspace{3mm} \\
    \bottomrule
    \end{tabular}
    \caption{Hyper-parameters of \textsc{LM-Lexicon-Dense} and \textsc{LM-Lexicon-MoE} training. \textsc{DS ZeRO-3} (left-hand table) denotes stage-3 ZeRO parallelism implemented by DeepSpeed \cite{rajbhandari2020zero}. \textsc{Naive PP} (right-hand table) denotes naive pipeline parallelism implemented by \HFLogo\ Hugging Face Transformers \cite{wolf2020transformers}.} 
    \label{tab:training-hyperparams}
\end{table*}
\begin{figure*}[ht]
  \centering
  \resizebox{1.0\linewidth}{!}{\includegraphics{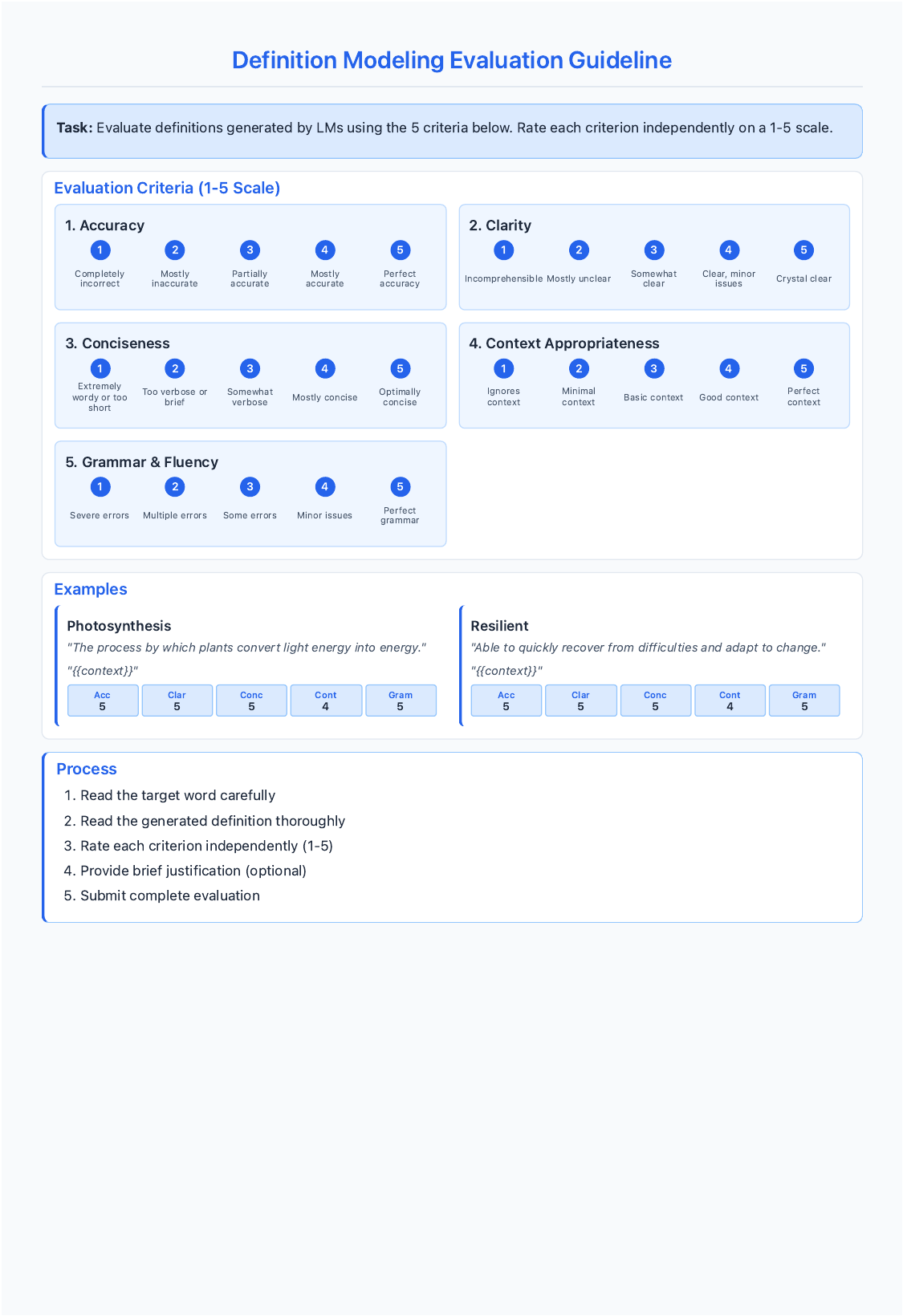}}
  \caption{Human evaluation guideline.}
  \label{fig:human_eval_guideline}
\end{figure*}


\end{document}